\PassOptionsToPackage{table,svgnames}{xcolor}
\documentclass[runningheads]{llncs}

\usepackage{eccv}
\usepackage{eccvabbrv}
\usepackage[utf8]{inputenc}
\usepackage[T1]{fontenc}
\usepackage{amsmath}
\usepackage{amssymb}
\usepackage{bm}
\usepackage{bbm}
\usepackage{graphicx}
\usepackage{wrapfig}
\usepackage{subcaption}
\usepackage{booktabs}
\usepackage{makecell}
\usepackage{array}
\usepackage{multirow}
\usepackage{diagbox}
\usepackage{arydshln}
\usepackage{longtable}
\usepackage{tabularx}
\usepackage{enumitem}
\usepackage{tcolorbox}
\usepackage{pifont}
\usepackage{bbding}
\usepackage{listings}
\usepackage[ruled,vlined]{algorithm2e}
\usepackage[accsupp]{axessibility}
\usepackage{url}

\usepackage[pagebackref,breaklinks,colorlinks]{hyperref}
\definecolor{MyBlue}{RGB}{0,172,238}
\definecolor{MyPink}{RGB}{237,0,139}
\definecolor{UpGreen}{RGB}{0,128,0}
\definecolor{DownRed}{RGB}{200,0,0}
\definecolor{HeaderBlue}{HTML}{E6E6E6}
\definecolor{fired}{RGB}{222,82,57}
\definecolor{fourier_green}{RGB}{0,128,0}
\definecolor{visual_yellow}{RGB}{191,144,0}
\definecolor{iceblue}{RGB}{33,102,200}
\definecolor{class}{HTML}{bb9726}
\definecolor{prompt}{HTML}{2978b5}
\definecolor{input}{HTML}{53b245}
\definecolor{mygray}{gray}{.9}
\definecolor{HeadGray}{HTML}{e6e6e6}
\definecolor{FRowGray}{RGB}{230,230,230}
\definecolor{headergray}{RGB}{227,226,233}
\definecolor{rowgray}{RGB}{242,242,242}
\definecolor{myblue}{HTML}{ECF4FF}
\definecolor{lightred}{RGB}{255,113,113}
\definecolor{lightblue}{RGB}{102,178,255}
\definecolor{mycyan}{HTML}{008092}

\hypersetup{
    colorlinks=true,
    linkcolor=red,
    citecolor=MyBlue,
    urlcolor=MyPink
}

\newcolumntype{C}[1]{>{\centering\arraybackslash}p{#1}}

\newcommand{\concat}{\mathbin{\|}}

\newcommand{\brdelta}[2]{%
    \makebox[2.60em][r]{#1}%
    \raisebox{-0.55ex}{\makebox[2.65em][l]{\scalebox{0.65}{%
                \ifdim #2pt > 0pt
                \textcolor{UpGreen}{\(\uparrow\)#2}%
                \else
                \textcolor{DownRed}{\(\downarrow\)#2}%
                \fi
    }}}%
}

\newcommand{\brdeltaF}[1]{%
    \makebox[2.60em][r]{#1}%
    \raisebox{-0.55ex}{\makebox[2.65em][l]{\scalebox{0.65}{\textcolor{gray}{(--)}}}}%
}

\newcommand{\red}[1]{\textcolor{red}{#1}}
\newcommand{\workdone}{\textsuperscript{*}}
\newcommand{\corr}{\textsuperscript{\ensuremath{\dagger}}}
\begin{document}
	
	\title{HIVE: Understanding Post-Hallucination Reasoning in Vision Language Models} 
	
	\titlerunning{HIVE}
	
	
	\author{
		Feng He\inst{1}\workdone \and
		Zhenting Wang\inst{2} \and
		Qifan Wang\inst{3} \and
		Qiang Guan\inst{4} \and
		Dongfang Liu\inst{1}\corr \and
		Ruixiang Tang\inst{2} \and
		Qiankun Li\inst{5}\corr
	}
	
	\authorrunning{F.~He et al.}
	
	\institute{
		Purdue University
		\and
		Rutgers University
		\and
		Meta AI
		\and
		Kent State University
		\and
		Imperial College London\\[0.5em]
		%	\email{hefengcs@gmail.com, zhenting.wang@rutgers.edu, wqfcr@meta.com, qguan@kent.edu, dfliu@purdue.edu, ruixiang.tang@rutgers.edu, q.li2@imperial.ac.uk}
	}
	\maketitle
	
	\begingroup
	\renewcommand{\thefootnote}{}
	\footnotetext{
		\textsuperscript{*}Work done while visiting Purdue University.\\
		\textsuperscript{\ensuremath{\dagger}}Corresponding authors.
	}
	\endgroup

	\begin{abstract}
		Hallucinations in vision--language models (VLMs) are commonly treated as semantic errors, yet they often arise from partial or ambiguous visual evidence. Prior work mainly focuses on detecting or suppressing hallucinations at generation time, leaving the subsequent reasoning stage largely unexplored. In this work, we study \emph{Post-Hallucination Reasoning} (PHR), the stage in which hallucinated semantics enter the model’s inference context and influence downstream predictions. To systematically investigate PHR, we introduce the \textbf{HIVE} (\textbf{H}allucination \textbf{I}nference and \textbf{V}erification \textbf{E}ngine), an evaluation infrastructure that enables controlled comparisons between faithful and hallucinated captions. Across nine tasks and nine models, we observe structured modality-dependent patterns: hallucinated captions often improve accuracy on vision--language tasks, while text-only tasks exhibit limited or unstable effects. Further analyses show that hallucinated cues broaden semantic coverage and reshape reasoning dynamics while preserving stable inference. These findings highlight that hallucinated semantics may influence downstream reasoning once they enter the model’s inference context. Understanding this post-hallucination stage is important for improving the reliability and interpretability of multimodal reasoning systems. Code is publicly available at \url{https://github.com/hefengcs/HIVE}.
		\keywords{VLMs \and Hallucination Analysis \and Multimodal Reasoning }
	\end{abstract}
	\section{Introduction}
	\label{sec:introduction}

	\begin{figure}[ht]
		\centering
		\includegraphics[width=1\linewidth]{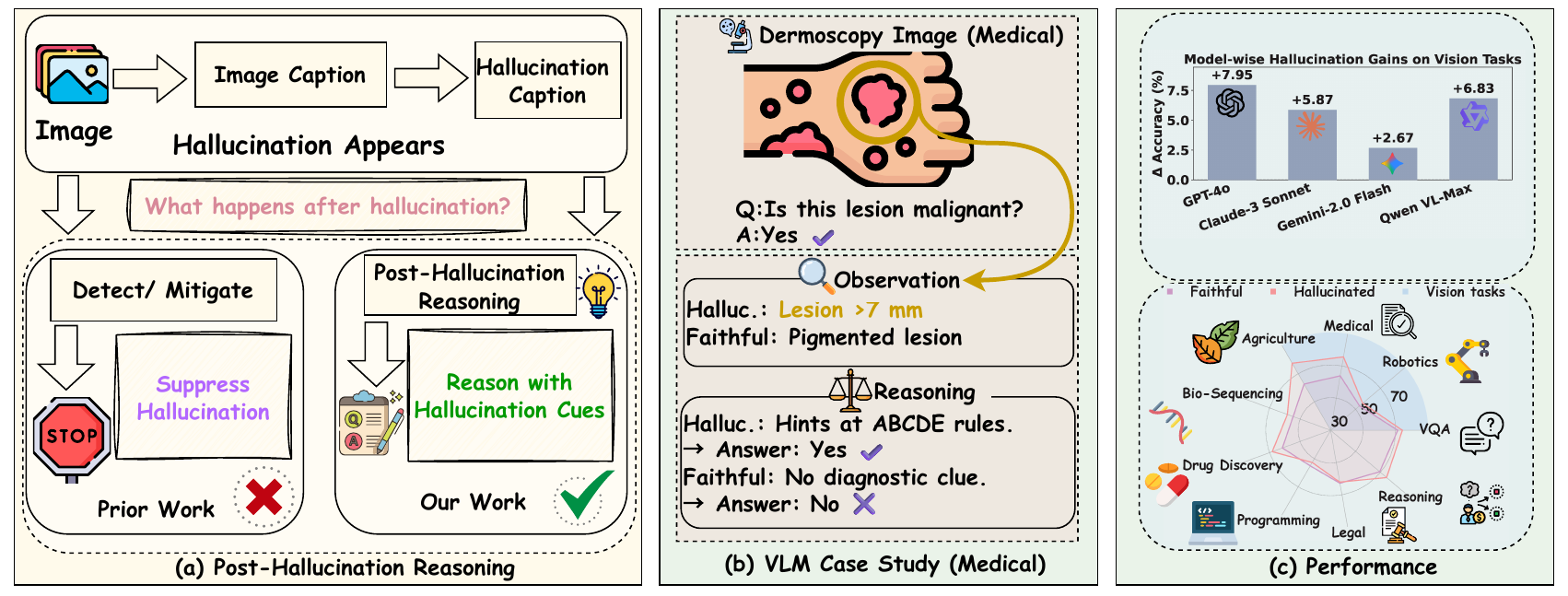}
		\caption{\textbf{Post–hallucination reasoning in VLMs.}
			\textbf{(a)} Prior work mainly focuses on detecting or suppressing hallucinations, whereas we study reasoning that occurs after hallucinations appear.
			\textbf{(b)} A medical example illustrates how hallucinated cues (e.g., lesion size) can alter the model’s observation and reasoning process.
			\textbf{(c)} Across models and vision tasks, hallucinations can systematically change task performance, revealing structured post--hallucination dynamics, especially in vision--language tasks.
		}
		\label{fig:overview}
		%\vspace{-20pt}
	\end{figure}

	\hspace*{2em}
	Vision--language models (VLMs) often operate under partial observability~\cite{rohrbach2018object,li2023evaluating,zhang2024vision,li2025survey,ma2024vision,hong2021vln}. Occlusion, low resolution, and motion blur frequently lead them to produce speculative semantic completions~\cite{krizhevsky2012imagenet,zellers2019recognition,wang2024mdpo}. According to community standards, such unverifiable content is considered \textit{hallucination}. However, hallucinations arise naturally from incomplete visual evidence rather than isolated system failures~\cite{ji2023survey,zhang2021hallucination,han2024few,han2023few,gong2024damro}. Consequently, most existing studies focus on detecting~\cite{li2023evaluating,manakul2023selfcheckgpt,luo2024hallucination,zhang2023enhancing,liu2025towards,su2024unsupervised,hu2024knowledge} or mitigating hallucinations~\cite{qiu2023detecting,wang2024mitigating,peng2025mitigating,zhou2025hademif,yin2025clearsight} at the moment they appear. Far less attention has been paid to what happens after hallucinated semantics enter the reasoning process of a VLM, leaving this post-hallucination stage largely unexplored. 
	
	Meanwhile, studies of LLMs show that chain-of-thought reasoning~\cite{wei2022chain,creswell2022faithful,xu2024faithful,li2025towards,parcalabescu2024measuring,lyu2023faithful,peng2025mitigating} does not strictly follow human causal logic: incorrect intermediate steps can still lead to correct conclusions. This observation suggests that reasoning may continue to evolve in latent space even when intermediate representations are imperfect or partially incorrect. By analogy, hallucinations in VLMs may also interact with downstream reasoning processes. However, whether such speculative cues help, hinder, or simply perturb reasoning remains unclear.
	
	We refer to this phenomenon as \textbf{Post-Hallucination Reasoning (PHR)}, which describes the reasoning behavior that emerges after hallucinated semantics become part of the model’s inference context and influence subsequent predictions. Despite the prevalence of hallucinations in multimodal systems, the structure and impact of PHR remain poorly understood. 
	As illustrated in Fig.~\ref{fig:overview}, once hallucinated semantics enter the model’s inference context, they may reshape the observation and reasoning process, ultimately influencing downstream predictions. Across diverse benchmarks, we observe a consistent pattern: hallucinated semantic cues can sometimes \emph{improve} downstream reasoning in vision-language tasks. This suggests that hallucinations may occasionally provide speculative semantic anchors that guide inference rather than merely degrading predictions. However, this effect is neither universal nor well understood. These observations raise two fundamental questions:
	\textbf{(1)} \textit{When does PHR improve downstream reasoning?}
	\textbf{(2)} \textit{Why does PHR lead to improved reasoning?}
	
	To systematically study these questions, we introduce \textbf{HIVE} 
	(\textbf{H}allucination \textbf{I}nference and \textbf{V}erification \textbf{E}ngine), an evaluation infrastructure for studying PHR. HIVE constructs paired faithful and hallucinated captions, verifies their validity, and enables controlled measurement of their influence on downstream predictions. By comparing raw inputs, faithful, and hallucinated captions under matched conditions, HIVE allows us to isolate the reasoning effects induced by hallucinated semantics and analyze the structure of PHR. Across 9 tasks and 9 models, we observe structured patterns: hallucinated semantic cues often improve performance on vision-language benchmarks, with gains varying systematically across model families and decoding settings.
	
	To understand the mechanisms underlying PHR, we analyze hallucination-induced reasoning dynamics at multiple levels.
	\textbf{(I)} Hallucinated captions broaden semantic coverage and widen embedding distributions.
	\textbf{(II)} They modulate reasoning entropy, with correct predictions correlating with higher entropy.
	
	These findings reveal \textbf{PHR} as an overlooked stage of multimodal inference, where hallucinated semantics can reshape and sometimes enhance downstream reasoning rather than merely introducing noise. Our contributions are:
	
	\begin{itemize}[topsep=1pt,itemsep=2pt,parsep=0pt,leftmargin=0pt,label={}]
		\item \textbf{\ding{182} Post-Hallucination Reasoning Phenomenon.} We identify and characterize \textbf{PHR}, a previously overlooked stage in which hallucinated semantics influence downstream reasoning in VLMs across diverse task settings. We observe structured patterns showing that hallucinated semantic cues can systematically reshape model predictions rather than merely introducing noise.
		
		\item \textbf{\ding{183} Evaluation Infrastructure for PHR.} 
		We present \textsc{HIVE}, an evaluation infrastructure designed to study 
		PHR through controlled semantic comparisons. 
		HIVE organizes caption generation, hallucination discrimination, and 
		downstream evaluation under matched configurations, enabling paired 
		comparisons between faithful and hallucinated captions and allowing us 
		to isolate and quantify the downstream impact across multiple models and task domains.
		
		\item \textbf{\ding{184} Mechanistic Analysis of PHR.} We analyze PHR across the input, reasoning-process, and output levels. Our results show that hallucinated captions broaden semantic coverage, induce distinctive distributional shifts, and modulate reasoning dynamics associated with successful predictions. These findings reveal how hallucinated semantics reshape reasoning trajectories in multimodal inference.
		
	\end{itemize}

	\vspace{-4mm}
	\section{Related Work}
	\vspace{-4mm}
	\hspace*{2em}
	\textbf{Hallucinations in VLMs: Detection and Mitigation.}
	Hallucination denotes content inconsistent with the given input. Hallucinations are widely observed in LLM and VLM outputs, appearing in tasks such as summarization \cite{zhao2020reducing} and open-domain QA \cite{sadat2023delucionqa}. Their presence undermines reliability in high-stakes domains, including healthcare \cite{lehman2021does,nori2023capabilities} and legal decision support \cite{guha2023legalbench,bendahman2025not}.  
	Two major threads dominate: detection and mitigation.  
	For detection, studies propose factuality metrics and benchmarks such as FactCC \cite{FactCC}, QAGS \cite{wang2020asking}, TruthfulQA \cite{lin2021truthfulqa}, and Q2 \cite{honovich2021q}, along with agreement- or internals-based judging \cite{manakul2023selfcheckgpt,du2024haloscope,su2024unsupervised}.  
	For mitigation, approaches include (I) instruction-level tuning \cite{zhang2024reflective,liu2023mitigating,yu2024hallucidoctor}, (II) constrained decoding \cite{lee2024delve,su2024unsupervised,choi2023kcts,mudgal2024controlled}, and (III) training or retrieval augmentation \cite{sennrich2023mitigating,manevich2024mitigating,lewis2020retrieval}.  
	These lines of work largely treat hallucinations as undesirable errors and focus on detecting or suppressing them, leaving the role of hallucinated semantics in subsequent reasoning largely unexplored.

	\textbf{Intermediate Reasoning and Latent Dynamics in LLMs.}
	Recent studies on language-only models reveal that a model’s reasoning process does not necessarily align with its explicit chain-of-thought text. Multiple works show that LLMs may generate incorrect, inconsistent, or partially fabricated intermediate steps while still producing correct final answers \cite{wei2022emergent,rajpurkar2018know}.  
	This phenomenon suggests that reasoning continues to evolve in the latent space and is not strictly determined by the faithfulness of intermediate text.  
	Further analyses indicate that LLMs can maintain stable internal inference even when intermediate steps contain hallucinated elements \cite{lanham2023measuring,wang2024chain,kadavath2022language}.  
	These findings imply that unfaithful or speculative content does not terminate reasoning; instead, the model proceeds to a subsequent reasoning stage that may still yield coherent conclusions.

	\textbf{Hallucinations in Multimodal Reasoning Models.}
	Recent work has begun to study hallucination effects in multimodal reasoning, showing that imperfect perceptual cues can influence reasoning outcomes in VLMs. 
	Studies such as More Thinking, Less Seeing~\cite{xu2026more}, More Thought, Less Accuracy~\cite{tian2026more}, MIRAGE~\cite{dong2026mirage}, and multimodal CoT analyses~\cite{ma2026understanding} indicate that hallucinations interact with reasoning dynamics beyond perception-level errors. 
	A recent survey further summarizes evaluation and detection efforts in this area~\cite{chen2026survey}. 
	However, these methods mainly analyze either intermediate reasoning traces or benchmark-level behavior, without explicitly studying how hallucinated semantics, once used as input context, affect downstream reasoning.
	
	\textbf{Different from existing methods}, we focus on a largely overlooked stage in VLM inference: \textbf{Post-Hallucination Reasoning (PHR)}, where hallucinated semantics, once generated and adopted by the model, may influence subsequent reasoning and task predictions. Prior work on VLM hallucinations focuses almost entirely on whether hallucinations occur, why they arise, and how to reduce them. Yet little attention has been paid to how hallucinated semantics affect the reasoning processes that follow. Consequently, the structure and impact of \emph{PHR} remain poorly understood, and the field lacks frameworks for controlled semantic interventions to study it. 
	
	\vspace{-4mm}
	\section{Evaluation Pipeline for PHR}
	\vspace{-3mm}
	
	\subsection{Problem Setup}
	\label{sec:problem}
	\vspace{-5pt}
	\hspace*{2em}
	HIVE is designed as an evaluation infrastructure for studying PHR in VLMs. It examines how hallucinated semantics, once generated and adopted by the model, influence subsequent reasoning and task predictions compared with faithful semantics. Rather than introducing a new model or training method, HIVE follows established community practices, including prompting-based caption generation and hallucination detection, and organizes them into a controlled semantic intervention framework for analyzing the causal effects of hallucinated semantics. To examine this question, we organize HIVE around three principles.
	
	\textbf{\ding{172} Fair comparison.} 
	We must ensure that the only difference between faithful and hallucinatory captions lies in the presence of hallucination itself. 
	Thus, captions are generated from a unified source with identical prompts, temperature, and token budget. 
	This design rules out confounds and allows us to focus on the difference in downstream performance across tasks and models, measured as the accuracy gap $\Delta(H-F)$ between hallucinated and faithful augmentations.
	
	\textbf{\ding{173} Task-agnostic hallucination generation.} 
	Hallucinations naturally arise when an LLM is asked to produce a free-form caption of any input, regardless of modality. Some captions remain faithful, while others introduce unverifiable or speculative elements. This inherent property enables us to adapt the paradigm seamlessly across diverse textual and multimodal tasks.
	
	\textbf{\ding{174} Reliable discrimination.} 
	Hallucination detection is inherently imperfect, and even human annotators may disagree.
	To enhance robustness, we adopt an ensemble of detectors that independently judge each caption, 
	and apply majority voting to obtain the final label. 
	This ensemble strategy ensures that the framework remains reliable under noisy classifiers. Formally, let $x \in \mathcal{X}$ denote an input with label $y \in \mathcal{Y}$, and
	$f: \mathcal{X} \times \mathcal{C} \to \mathcal{Y}$ a task-specific model. Each input can
	be paired with a faithful caption $\bm{C}_F$ or a hallucinatory caption $\bm{C}_H$. Such a triplet design ensures that hallucination is the only varying factor across conditions, thus enabling a controlled and interpretable evaluation. Formally, we define these conditions:
	% \begin{equation}
		% y_{\textsc{raw}} = \bm{f}(x), \quad
		% y_{F} = \bm{f}(x \concat \bm{C}_{F}), \quad
		% y_{H} = \bm{f}(x \concat \bm{C}_{H}),
		% \end{equation}
	\begin{equation}
		y_{\textsc{raw}} = \underbrace{\bm{f}(x)}_{\text{Raw}},\quad
		y_{F}           = \underbrace{\bm{f}(x \concat \bm{C}_{F})}_{\text{+ Faithful}},\quad
		y_{H}           = \underbrace{\bm{f}(x \concat \bm{C}_{H})}_{\text{+ Hallucinatory}},
	\end{equation}
	where $\concat$ denotes concatenation with the task instruction.
	Given an evaluation metric $\mathcal{L}(\hat{y}, y)$ (instantiated as accuracy in our experiments),
	we quantify the hallucination effect as
	% 	\begin{equation}
		% 	\Delta(H-F) = \mathbb{E}_{(x,y)\sim \mathcal{D}}
		% 	\Big[ \ell(f(x \,\Vert\, C_H), y) - \ell(f(x \,\Vert\, C_F), y) \Big].
		% 	\end{equation}
	%     \begin{equation}
		\begin{equation}
			\bm{\Delta}(H{-}F) \;=\; 
			\mathbb{E}_{(x,y)\sim \bm{\mathcal{D}}}
			\Big[\, \mathcal{L}\!\big(\bm{f}(x \concat \bm{C}_H),\, y\big)
			- \mathcal{L}\!\big(\bm{f}(x \concat \bm{C}_F),\, y\big) \,\Big].
		\end{equation}
		This paired comparison isolates hallucination as a controlled experimental variable and enables apples-to-apples analysis of \textbf{PHR} across models and tasks.
		
		\subsection{HIVE Evaluation Pipeline}
		\label{sec:hive-workflow}
		\hspace*{2em}
		%	As described in Algorithm~\ref{alg:hive}, the HIVE framework consists of three modules:
		The HIVE evaluation pipeline consists of three modules: 
		(I) \textbf{Caption Generator}, (II) \textbf{Caption Discriminator}, and (III) \textbf{Task Solver}. 
		Given an input instance, the pipeline proceeds through these components to ensure consistent and controlled evaluation across tasks.
		%	\begin{algorithm}[t]
			%		\caption{HIVE pipeline from semantic contrastive view: Faithful (F) vs. Hallucinated (H).}
			%		\label{alg:hive}
			%		
			%		\KwIn{Dataset $\bm{\mathcal{D}}$, task model $\bm{f}$, generator $G$, discriminator $D$}
			%		\KwOut{Performance difference $\bm{\Delta}(H{-}F)$}
			%		
			%		\ForEach{$(x,y) \in \bm{\mathcal{D}}$}{
				%			Generate candidate captions $C(x) = \{c_1, \dots, c_N\}$ via $G$ \textcolor{mycyan}{// Expose semantic diversity} \\
				%			Classify each $c_i$ as Faithful or Hallucinatory using $D$ \textcolor{mycyan}{// Partition as F/H} \\
				%			\If{a contrasted pair $\langle \bm{C}_F, \bm{C}_H \rangle$ exists}{
					%				$y_{\textsc{raw}} \gets \bm{f}(x)$ \textcolor{mycyan}{// Setting for raw input} \\
					%				$y_F \gets \bm{f}(x \concat \bm{C}_F)$ \textcolor{mycyan}{// Setting for faithful caption} \\
					%				$y_H \gets \bm{f}(x \concat \bm{C}_H)$ \textcolor{mycyan}{// Setting for hallucinatory caption} \\
					%				Record $\mathcal{L}(y_H,y)$ and $\mathcal{L}(y_F,y)$\;
					%			}
				%		}
			%		Compute $\bm{\Delta}(H{-}F) = \mathbb{E}_{(x,y)\sim \bm{\mathcal{D}}}
			%		[\,\mathcal{L}(y_H,y) - \mathcal{L}(y_F,y)\,]$ \textcolor{mycyan}{// Hallucination effect measure} \\
			%	\end{algorithm}
		The Caption Generator first takes the raw input (text, image, or structured record) 
		and produces a set of candidate captions under a unified, task-agnostic prompt, 
		which may include both faithful and hallucinatory semantics. 
		The Caption Discriminator then evaluates these candidates and classifies each as 
		either faithful ($\bm{C}_F$) or hallucinatory ($\bm{C}_H$); only contrasted pairs 
		$\langle \bm{C}_F, \bm{C}_H \rangle$ with majority agreement among detectors are retained. 
		Finally, the Task Solver integrates the original input $x$, one of the paired captions, 
		and a task-specific instruction to produce the final prediction $y$. 
		This design yields three controlled conditions Raw ($y_{\textsc{raw}}=\bm{f}(x)$), 
		+Faithful ($y_F=\bm{f}(x \concat \bm{C}_F)$), and +Hallucinatory ($y_H=\bm{f}(x \concat \bm{C}_H)$).
		
		\textbf{Caption Generator.} The Caption Generator aims to produce diverse semantic candidates that may include both faithful and hallucinatory variants. All captions are generated from a unified source using the same prompt, temperature, and token budget, 
		ensuring that decoding hyper-parameters cannot confound attribution. 
		Given an input $x$, the generator outputs $N$ natural-language captions describing it. 
		Due to the inherent stochasticity of LLMs, some captions remain faithful while others introduce speculative content, 
		which later enables the construction of contrasted F/H pairs by the discriminator. 
		This design guarantees that both F and H captions originate from an identical generation process, 
		providing a controlled entry point for subsequent evaluation. 
		
		\textbf{Caption Discriminator.} Given the candidate captions, the Caption Discriminator determines whether each caption is faithful ($C_F$) or hallucinatory ($C_H$). Since hallucination detection is inherently noisy, we adopt an ensemble of multiple detectors, each providing an independent judgment. Final labels are decided via majority voting, which significantly improves robustness under noisy or imperfect classifiers. We further verify the reliability of the hallucination discriminator through dedicated control experiments (\red{Section}~\ref{sec:Validating}). Detailed implementation specifics of the individual detectors and ensemble configuration are provided in Appendix \red{S7} for full clarity and reproducibility.
		
		% \textbf{(III) Task Solver.} The solver is not a novel model but a controlled interface to isolate the effect of captions on downstream predictions. We measure the isolated impact of caption faithfulness by contrasting predictions under three conditions $C\in\{\varnothing,C_{\mathrm{F}},C_{\mathrm{H}}\}$.
		% We use a unified prompt builder that concatenates three parts:
		% \begin{equation}
			% 	\Phi(\mathcal{I}_{\text{task}},x,C)\;=\;
			% 	\underbrace{\text{\textsc{Instruction} }\,\mathcal{I}_{\text{task}}}_{\text{fixed}}
			% 	\;\Vert\;
			% 	\underbrace{\text{\textsc{Serialized Input} }\,\sigma(x)}_{\text{image/table/text}}
			% 	\;\Vert\;
			% 	\underbrace{\text{\textsc{Caption} }\,s(C)}_{\text{style- and length-controlled}},
			% 	\label{eq:phi}
			% \end{equation}
		% where $\Vert$ denotes newline separation, $\sigma(\cdot)$ serializes $x$, and $s(\cdot)$ is a deterministic normalizer.
		% Prompt templates for all benchmarks are comprehensively listed in \S\ref{app:Prompt_Templates} 
		%    to ensure clarity.
		
		\textbf{Task Solver.} 
		The solver is not a novel model but a controlled interface to isolate the effect of captions on downstream predictions under identical model and decoding settings. 
		We measure the isolated impact of caption faithfulness by contrasting predictions under three conditions $C \in \{\textsc{raw}, \bm{C}_{F}, \bm{C}_{H}\}$. 
		We use a unified prompt builder that concatenates three parts:
		\begin{equation}
			{\small
				\Phi(\mathcal{I}_{\text{task}},x,C)=
				\textsc{Instruction}\,\mathcal{I}_{\text{task}}
				\mathbin{\|}
				\textsc{Serialized Input}\,\sigma(x)
				\mathbin{\|}
				\textsc{Seed}\,s(C)
			}
		\end{equation}
		HIVE frames hallucination as a controlled
		semantic intervention, enabling systematic study
		of how speculative cues influence downstream
		reasoning in multimodal models.
		Prompt templates are listed in Appendix \red{S2}.
		
		\vspace{-3mm}
		\section{Experiments}
		\vspace{-2mm}
		\subsection{When Does PHR Help?}
		\label{sec:main}
		\hspace*{2em}
		We begin by characterizing how hallucinated semantics affect downstream predictions across a diverse set of models and tasks. Using the HIVE evaluation infrastructure, we systematically analyze PHR across 9 tasks and 9 models spanning both textual and multimodal scenarios. Each model is evaluated under two input conditions: faithful and hallucinated. \autoref{tab:text-main} reports results on text-only benchmarks, while \autoref{tab:multi-main} summarizes vision–language tasks. Further dataset and model details are provided in Appendix~\red{S6}.
		
		\textbf{ \ding{182} \textbf{LLMs on text-only tasks: limited or unstable benefits.}} On textual tasks, hallucinations provide little or no benefit. Across multiple LLM families, performance under hallucinated inputs is often similar to or lower than the faithful setting, with only occasional isolated improvements (see \autoref{tab:text-main}). The overall trends are small and inconsistent, indicating that hallucinated semantics are hard to exploit in rule-based or symbolically constrained tasks.
		
		\textbf{ \ding{183} \textbf{VLMs on vision–language tasks: strong and consistent gains.}} On vision--language benchmarks, hallucinated inputs lead to clear and substantial improvements across models (see \autoref{tab:multi-main}). GPT-4o, Gemini-2.0-Flash, and Qwen-VL-Max generally show gains, with occasional drops on specific datasets such as GQA in certain settings. Unlike the text-only setting, where hallucination effects are small and unstable, VLMs reliably benefit from hallucinated semantics across multiple tasks and model families. 
		
		These results suggest that hallucinated captions can enrich visual grounding by introducing additional perceptual cues that support downstream reasoning. Vision–language tasks often involve partial observability, where important cues may be missing or ambiguous. In such settings, hallucinated captions can serve as speculative hypotheses that expand the model’s hypothesis space and provide additional semantic anchors for reasoning. 
		
		\textbf{Model-wise variation.} While this overall pattern holds across VLMs, the magnitude of PHR gains varies by model. For example, GPT-4o exhibits moderate but stable improvements, whereas Gemini-2.0-Flash and Qwen-VL-Max often show larger gains on perception-heavy tasks. This variation likely reflects differences in visual grounding strength and reasoning strategies across models: models with stronger perceptual grounding can better exploit speculative semantic cues introduced by hallucinated captions, while others benefit less.

		\begin{table}[t]
			\centering
			\caption{\textbf{Faithful (F) vs. Hallucinated (H) path accuracy on text-only tasks.}
				Cells show accuracy (\%). $\Delta(H-F)$ denotes the percentage-point difference between the hallucinated and faithful paths
				(\textcolor{UpGreen}{$\uparrow$} gain, \textcolor{DownRed}{$\downarrow$} drop).}
			\label{tab:text-main}
			
			{\scriptsize
				\setlength{\tabcolsep}{0pt}
				\renewcommand{\arraystretch}{0.96}
				\begin{tabular*}{\textwidth}{@{\extracolsep{\fill}}lc*{7}{c}@{}}
					\toprule
					\textbf{Dataset} & \textbf{P.} &
					\textbf{GPT-4o} &
					\textbf{GPT-3.5} &
					\textbf{\makecell[c]{Claude-3\\Sonnet}} &
					\textbf{\makecell[c]{DeepSeek\\v3}} &
					\textbf{\makecell[c]{Mistral\\Large}} &
					\textbf{O3} &
					\textbf{\makecell[c]{DeepSeek\\R1}} \\
					\midrule
					
					\rowcolor{FRowGray}
					AntiCP2 & F &
					\brdeltaF{54.59} & \brdeltaF{43.63} & \brdeltaF{46.84} &
					\brdeltaF{52.19} & \brdeltaF{56.69} & \brdeltaF{53.95} & \brdeltaF{49.87} \\
					& H &
					\brdelta{58.35}{+3.76} & \brdelta{47.76}{+4.13} & \brdelta{48.21}{+1.37} &
					\brdelta{45.01}{-7.18} & \brdelta{57.84}{+1.15} & \brdelta{50.17}{-3.78} &
					\brdelta{46.42}{-3.45} \\
					\midrule
					
					\rowcolor{FRowGray}
					BBBP & F &
					\brdeltaF{61.67} & \brdeltaF{60.75} & \brdeltaF{64.07} &
					\brdeltaF{61.60} & \brdeltaF{59.41} & \brdeltaF{73.27} & \brdeltaF{70.53} \\
					& H &
					\brdelta{68.33}{+6.66} & \brdelta{57.53}{-3.22} & \brdelta{64.95}{+0.88} &
					\brdelta{55.56}{-6.04} & \brdelta{59.65}{+0.24} & \brdelta{59.47}{-13.80} &
					\brdelta{58.88}{-11.65} \\
					\midrule
					
					\rowcolor{FRowGray}
					CodeXGLUE & F &
					\brdeltaF{55.15} & \brdeltaF{58.90} & \brdeltaF{49.25} &
					\brdeltaF{49.46} & \brdeltaF{50.11} & \brdeltaF{45.10} & \brdeltaF{51.54} \\
					& H &
					\brdelta{52.75}{-2.40} & \brdelta{57.40}{-1.50} & \brdelta{53.40}{+4.15} &
					\brdelta{50.13}{+0.67} & \brdelta{56.40}{+6.29} & \brdelta{46.06}{+0.96} &
					\brdelta{48.95}{-2.59} \\
					\midrule
					
					\rowcolor{FRowGray}
					SARA v3 & F &
					\brdeltaF{62.93} & \brdeltaF{52.28} & \brdeltaF{62.07} &
					\brdeltaF{58.10} & \brdeltaF{59.14} & \brdeltaF{65.52} & \brdeltaF{58.62} \\
					& H &
					\brdelta{62.24}{-0.69} & \brdelta{54.83}{+2.55} & \brdelta{63.97}{+1.90} &
					\brdelta{58.96}{+0.86} & \brdelta{60.34}{+1.20} & \brdelta{62.07}{-3.45} &
					\brdelta{54.48}{-4.14} \\
					\midrule
					
					\rowcolor{FRowGray}
					ProofWriter & F &
					\brdeltaF{69.49} & \brdeltaF{66.55} & \brdeltaF{76.03} &
					\brdeltaF{85.17} & \brdeltaF{70.86} & \brdeltaF{97.76} & \brdeltaF{89.31} \\
					& H &
					\brdelta{75.00}{+5.51} & \brdelta{66.21}{-0.34} & \brdelta{76.73}{+0.70} &
					\brdelta{85.86}{+0.69} & \brdelta{68.62}{-2.24} & \brdelta{98.45}{+0.69} &
					\brdelta{92.93}{+3.62} \\
					\bottomrule
				\end{tabular*}
			}
			%\vspace{-4pt}
		\end{table}
		
\begin{table}[t]
	\centering
	\caption{\textbf{Faithful (F) vs. Hallucinated (H) accuracy on vision-language tasks.}
		Cells show mean accuracy (\%). $\Delta$ denotes $H-F$ in percentage points and is shown inline at bottom-right.}
	\label{tab:multi-main}
	
	{\scriptsize
		\setlength{\tabcolsep}{2.2pt}
		\renewcommand{\arraystretch}{0.96}
		\begin{tabularx}{0.88\textwidth}{@{}l c *{4}{>{\centering\arraybackslash}X}@{}}
			\toprule
			\textbf{Dataset} & \textbf{P.} &
			\textbf{GPT-4o} &
			\textbf{\makecell[c]{Claude 3\\Sonnet}} &
			\textbf{\makecell[c]{Gemini 2.0\\Flash}} &
			\textbf{\makecell[c]{Qwen\\VL-Max}} \\
			\midrule
			
			\rowcolor{FRowGray}
			GQA & F &
			\brdeltaF{71.36} & \brdeltaF{62.02} & \brdeltaF{75.23} & \brdeltaF{69.88} \\
			& H &
			\brdelta{74.22}{+2.86} & \brdelta{61.03}{-0.99} &
			\brdelta{75.69}{+0.46} & \brdelta{67.90}{-1.98} \\
			\midrule
			
			\rowcolor{FRowGray}
			Dex-Net & F &
			\brdeltaF{53.25} & \brdeltaF{50.29} & \brdeltaF{49.88} & \brdeltaF{51.54} \\
			& H &
			\brdelta{55.76}{+2.51} & \brdelta{50.71}{+0.42} &
			\brdelta{51.15}{+1.27} & \brdelta{54.12}{+2.58} \\
			\midrule
			
			\rowcolor{FRowGray}
			ISIC & F &
			\brdeltaF{63.88} & \brdeltaF{54.19} & \brdeltaF{67.23} & \brdeltaF{58.71} \\
			& H &
			\brdelta{75.64}{+11.76} & \brdelta{61.02}{+6.83} &
			\brdelta{67.90}{+0.67} & \brdelta{75.62}{+16.91} \\
			\midrule
			
			\rowcolor{FRowGray}
			PlantVillage & F &
			\brdeltaF{62.73} & \brdeltaF{55.28} & \brdeltaF{62.71} & \brdeltaF{67.84} \\
			& H &
			\brdelta{77.41}{+14.68} & \brdelta{72.50}{+17.22} &
			\brdelta{70.98}{+8.27} & \brdelta{77.66}{+9.82} \\
			\bottomrule
		\end{tabularx}
	}
	%\vspace{-4pt}
\end{table}
		
		\begin{figure}[t]
			\centering
			\includegraphics[width=1\textwidth]{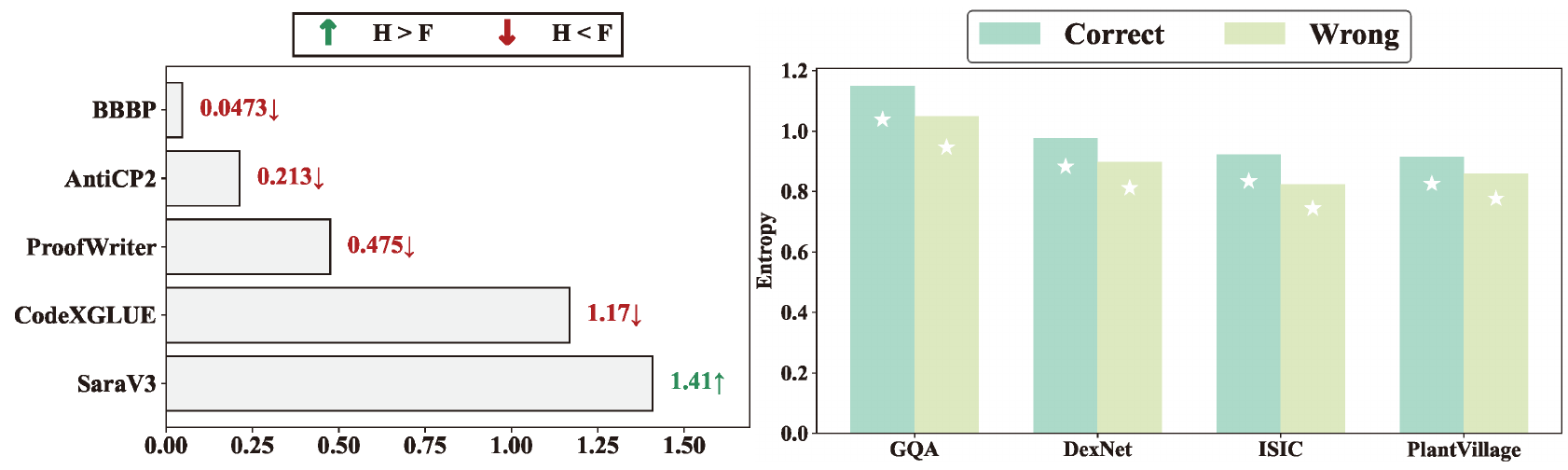}
			\caption{
				\textbf{Process- and output-level analysis of hallucination effects.} 
				\textbf{Left}: reasoning chain embeddings show that hallucinated prompts reshape inference trajectories with task-dependent differences ($p < 0.05$), diversifying reasoning. 
				\textbf{Right}: caption entropy analysis shows that correct predictions exhibit higher entropy than incorrect ones, indicating expanded semantic coverage. Stars indicate $p < 0.05$.
			}
			\label{fig:Entropy}
			\vspace{-10pt}
		\end{figure}
		
		\vspace{-4mm}
		\subsection{Validating the Hallucination Discriminator}
		\label{sec:Validating}
		\hspace*{2em}
		\textbf{Reliability on hallucination benchmarks.} To assess the reliability of HIVE's hallucination discriminator, we evaluate it in two complementary settings.
		First, we test the discriminator on the TruthfulQA benchmark, which is widely used to probe hallucination behavior in language models.
		The discriminator achieves an accuracy of \textbf{81.76\%} on this benchmark. To approximate real-world cross-domain usage, we construct a curated dataset of 180 captions by sampling 20 captions from each of nine tasks.
		These captions are manually annotated as either \emph{hallucinated} or \emph{faithful}.
		Evaluating the discriminator on this human-labeled dataset yields an accuracy of \textbf{83.72\%}. These results indicate that the discriminator generalizes beyond a single benchmark and can reliably separate hallucinated from faithful captions, providing a stable basis for the paired comparisons used throughout our experiments.
		
		\textbf{Random-checker control.} To rule out improvements arising from arbitrary filtering, we replace the hallucination
		discriminator with a \emph{random} checker that accepts captions without factual
		assessment. All decoding controls remain fixed to ensure a fair comparison.
		As shown in \autoref{tab:random-checker}, random filtering yields only negligible
		performance changes (0.17--1.23\%) and no statistically significant gains across
		datasets (all $p>0.14$). This result indicates that the improvements reported in
		\red{Section} \ref{sec:main} are unlikely to arise from chance filtering effects.

		\begin{table*}[htbp]
			\centering
			\footnotesize
			\setlength{\tabcolsep}{4pt}
			\renewcommand{\arraystretch}{1.0}
			
			\caption{\textbf{Random checker ablation.}
				$\Delta$ denotes the absolute accuracy difference (H--F).
				$p$ from two-sided paired $t$-tests; (n.s.) = not significant at $p{<}0.05$.}
			
			\begin{tabular}{lcccc}
				\Xhline{1.2pt}
				\rowcolor{CadetBlue!20}
				\textbf{Dataset} & \textbf{Faithful (F)} & \textbf{H + Random (H)} & $\boldsymbol{\Delta}$ (H--F) & \textbf{$p$} \\
				\Xhline{1.0pt}
				\rowcolor{gray!10}
				AntiCP2 & $0.5015 \pm 0.0124$ & $0.5114 \pm 0.0200$ & $+0.0100$ & $0.526$ (n.s.) \\
				% \rowcolor{gray!10}
				PlantVillage & $0.7358 \pm 0.0189$ & $0.7481 \pm 0.0205$ & $+0.0123$ & $0.354$ (n.s.) \\
				\rowcolor{gray!10}
				Dex-Net & $0.4950 \pm 0.0068$ & $0.4967 \pm 0.0062$ & $+0.0017$ & $0.757$ (n.s.) \\
				% \rowcolor{gray!10}
				ISIC & $0.5763 \pm 0.0067$ & $0.5805 \pm 0.0042$ & $+0.0043$ & $0.142$ (n.s.) \\
				\Xhline{1.2pt}
			\end{tabular}
			%\vspace{-4mm}
			\label{tab:random-checker}
		\end{table*}

		\textbf{Prompt robustness.}
		To rule out the possibility that the observed gains arise from prompt sensitivity,
		we repeat the experiments using semantically equivalent prompt paraphrases.
		Specifically, we evaluate three prompts: the original prompt (P0) and two paraphrased variants (P1 and P2). The results are summarized in \autoref{tab:prompt-robustness}. Across datasets and models, hallucination-augmented inputs consistently outperform
		faithful inputs under all prompt variants. Although absolute accuracies vary slightly
		due to known prompt sensitivity in LLMs, the direction of the effect remains stable.
		This suggests that the observed improvements are not tied to a specific prompt formulation.
		
		\begin{table*}[htbp]
			\centering
			\footnotesize
			\setlength{\tabcolsep}{3pt}
			\renewcommand{\arraystretch}{1.0}
			\vspace{-4mm}
			\caption{\textbf{Prompt robustness across paraphrased prompts.}
				P0 denotes the original prompt, while P1 and P2 are semantically equivalent paraphrases.
				$\Delta$ reports the accuracy improvement from hallucination injection (H--F).}

			\setlength{\tabcolsep}{3pt}
			
			\begin{tabular}{lcccc}
				\Xhline{1.2pt}
				\rowcolor{CadetBlue!20}
				\textbf{Dataset / Model} & \textbf{P0 $\Delta$ (\%)} & \textbf{P1 $\Delta$ (\%)} & \textbf{P2 $\Delta$ (\%)} & \textbf{Consistency} \\
				\Xhline{1.0pt}
				\rowcolor{gray!10}
				PlantVillage / GPT-4o & +14.68 & +28.00 & +9.89 & \textbf{\checkmark} \\
				% \rowcolor{gray!10}
				PlantVillage / Claude 3 Sonnet & +17.22 & +16.42 & +12.84 & \textbf{\checkmark} \\
				\rowcolor{gray!10}
				ISIC / GPT-4o & +11.76 & +4.55 & +5.00 & \textbf{\checkmark} \\
				% \rowcolor{gray!10}
				ISIC / Claude 3 Sonnet & +6.83 & +4.54 & +4.77 & \textbf{\checkmark} \\
				\Xhline{1.2pt}
			\end{tabular}
			\label{tab:prompt-robustness}
			%\vspace{-4mm}
		\end{table*}
		
		\textbf{Token-level ablation.} We conduct a token-level ablation study to test whether hallucinated content is
		\emph{necessary} for successful predictions. Specifically, we first identify the
		subset of samples that are solved only when hallucinated captions are provided,
		where the hallucinated path succeeds while the faithful path fails.
		Within these captions, we locate hallucinated tokens that serve as \emph{core
			evidence} in the model's reasoning process. We then mask these tokens using a neutral placeholder and re-evaluate the model
		on the same samples. As reported in \autoref{tab:abl-h-only}, we evaluate a subset of samples where the hallucinated path succeeds while the faithful path fails. For this subset, \emph{H-before} is near-perfect by construction, and post-ablation accuracy (\emph{H-after}) drops substantially across all four datasets after masking hallucinated evidence tokens. This result indicates that key hallucinated tokens act as
		informative cues rather than redundant noise, and that the model relies on
		them to reach correct predictions.

			\begin{table}[t]
				\centering
				\footnotesize
				\setlength{\tabcolsep}{6pt}
				\renewcommand{\arraystretch}{1.0}
				
				\caption{\textbf{Accuracy after hallucination removal across domains.}
					H-after Acc. denotes post-ablation accuracy after masking hallucinated evidence tokens.}
				\label{tab:abl-h-only}
				
				\begin{tabular}{llll}
					\Xhline{1.2pt}
					\rowcolor{CadetBlue!20}
					\textbf{Dataset} & \textbf{Domain} & \textbf{Task} & \textbf{H-after Acc.} \\
					\Xhline{1.0pt}
					\rowcolor{gray!10}
					AntiCP2 & Biomedicine & Peptide cls. & $0.244 \pm 0.085$ \\
					PlantVillage & Agriculture & Disease recog. & $0.700 \pm 0.111$ \\
					\rowcolor{gray!10}
					Dex-Net & Robotics & Grasping pred. & $0.364 \pm 0.061$ \\
					ISIC & Dermatology & Lesion diagnosis & $0.380 \pm 0.062$ \\
					\Xhline{1.2pt}
				\end{tabular}
			\end{table}

				\vspace{-4mm}
				\subsection{Reasoning Convergence Analysis}
\begin{figure}[t]
					\centering
					\includegraphics[width=1.0\textwidth]{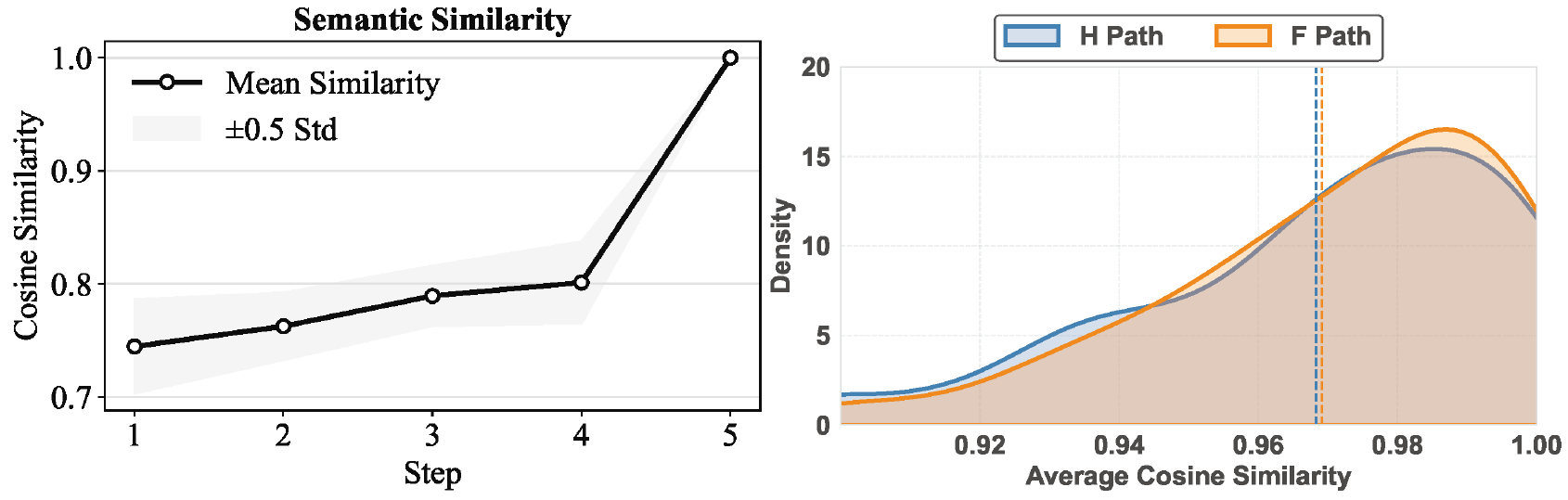}
					\caption{\textbf{Inter-chain stability on the \textsc{PlantVillage} dataset.}
						\textbf{Left}: Step-wise cosine similarity shows reasoning chains progressively converge during inference.
						\textbf{Right}: Hallucinated (H) and faithful (F) captions exhibit highly overlapping similarity distributions,
						indicating that hallucinations preserve stability across sampling runs.
					}
					
					\label{fig:Chain_PlantVillage}
					\vspace{-5mm} 
				\end{figure}

				\vspace{-2mm}
				\hspace*{2em}
				To assess whether the semantic shifts introduced by hallucinated captions 
				affect the stability of model inference, we analyze reasoning convergence at 
				two complementary levels: 
				(I) \textbf{Intra-chain convergence}, which measures whether intermediate 
				reasoning steps progressively align with the final conclusion under 
				hallucinated inputs (\autoref{fig:Chain_PlantVillage} left); and 
				(II) \textbf{Inter-chain consistency}, which evaluates whether multiple 
				reasoning paths sampled from the same input converge to semantically similar 
				trajectories across different sampling seeds (\autoref{fig:Chain_PlantVillage} 
				right). 
				These analyses provide a fine-grained view of how hallucinations 
				affect convergence both within and across reasoning chains, enabling us to 
				determine whether they destabilize or preserve the model’s inference process. 
				Further implementation details are provided in Appendix~\red{S9}.
				
				\textbf{\ding{184} Hallucinations do not disrupt intra-chain convergence.}
				From \autoref{fig:Chain_PlantVillage} (left), step-to-final semantic similarity 
				steadily increases, with intermediate steps progressively aligning with the 
				final conclusion. This trajectory indicates that reasoning chains naturally 
				converge as inference unfolds rather than drifting from the target answer. 
				Moreover, the narrowing variance band suggests that this convergence pattern 
				remains stable across runs and datasets.
				
				\textbf{\ding{185} Reasoning chains exhibit strong inter-chain consistency.}  
				From \autoref{fig:Chain_PlantVillage} (right), we observe that reasoning paths generated under hallucinated (H) and faithful (F) captions both achieve very high pairwise similarity across multiple sampling runs (means $\approx 0.97$).  
				The two distributions nearly overlap, and statistical tests confirm no significant difference between them ($p > 0.6$).  
				This indicates that, regardless of whether captions contain hallucinations, the model converges to consistent reasoning trajectories across chains. Rather than diverging into unstable alternatives, multiple sampled paths remain semantically aligned, underscoring the robustness of the model’s inference.
				\begin{tcolorbox}
					\textbf{Takeaway \ding{182}.} Reasoning chains with hallucinated captions remain stable: intermediate steps consistently converge to the final answer and multiple paths sampled stay highly aligned. This stability highlights that hallucinations can support reliable and reproducible inference.
				\end{tcolorbox}	
				
				\vspace{-4mm}
				\subsection{Why PHR Helps}
				\vspace{-3mm}
				%\begin{wrapfigure}{r}{0.5\textwidth}
				%	\centering
				%	\includegraphics[width=0.5\textwidth]{figures/zu3}
				%	\caption{\textbf{Distribution of caption embeddings (Faithful (F) vs. Hallucinated (H)).} 
					%		Hallucinated inputs exhibit wider semantic spread and longer tails. 
					%		Stars indicate $p < 0.01$.}
				%	\label{fig:distribution}
				%	%\vspace{-15pt}
				%\end{wrapfigure}
				\begin{figure}[t]
					\centering
					\includegraphics[width=0.7\linewidth]{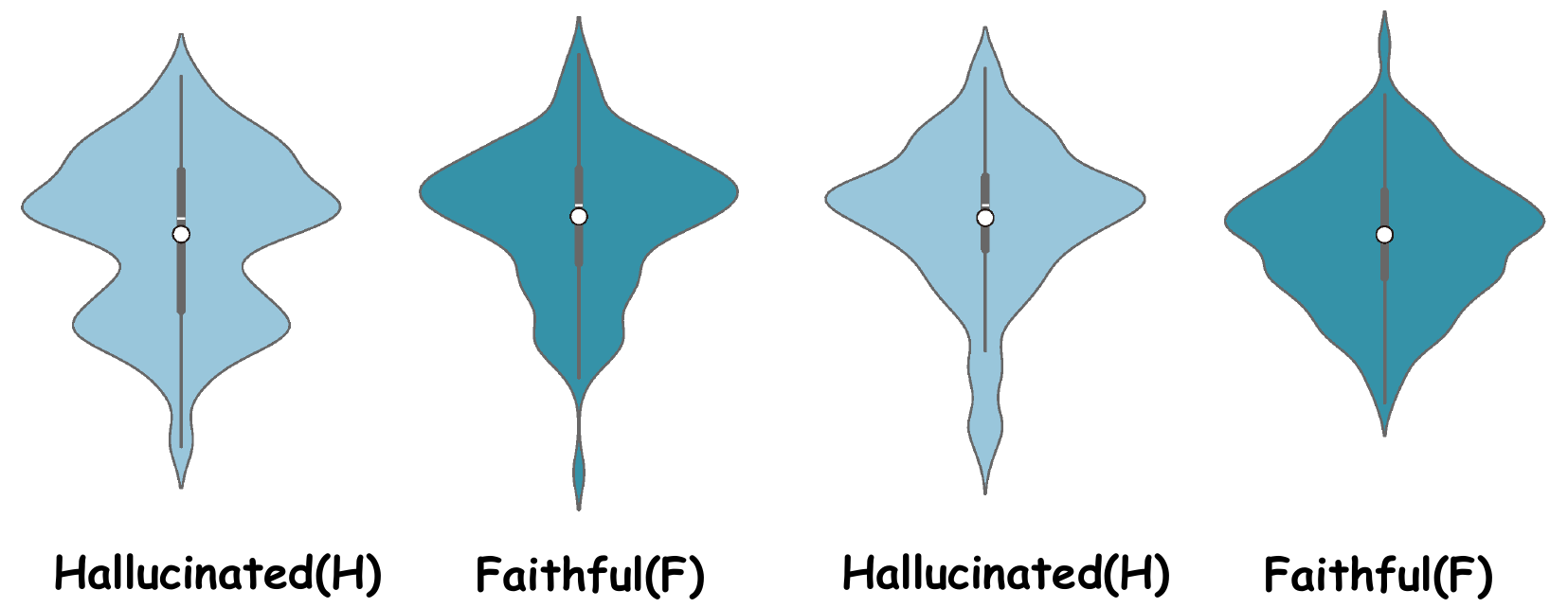}
					\caption{\textbf{Distribution of caption embeddings (Faithful (F) vs. Hallucinated (H)).}
						Hallucinated inputs exhibit wider semantic spread and longer tails.
					}
					\label{fig:distribution}
					%\vspace{-6mm}
				\end{figure}

				\hspace*{2em}
				To explain the strong and consistent improvements observed on 
				vision--language tasks (see Section \ref{sec:main}), we analyze why hallucinated captions 
				provide useful semantic signals for VLMs. Our analysis focuses on the 
				beneficial case and examines hallucination effects at three 
				complementary levels: \textbf{(I) Input-level shifts}. Hallucinated captions substantially reshape the semantic inputs provided 
				to the model, exhibiting broader distributional spread and lower similarity 
				to faithful captions (\autoref{fig:distribution}). This indicates that 
				hallucinations introduce meaningful semantic variation rather than redundant 
				noise, enriching the model’s visual grounding.
				\textbf{(II) Process-level modulation}. 
				Reasoning-chain entropy analysis (\autoref{fig:Entropy} left) shows that 
				hallucinations modulate inference dynamics in a task-dependent manner. 
				They reduce trajectory entropy on reasoning-heavy tasks (promoting 
				convergence and stability), while increasing entropy on structurally 
				open-ended tasks (supporting exploration), indicating active reshaping 
				of the model’s reasoning process.
				\textbf{(III) Output-level diversity}. 
				Correct predictions consistently exhibit higher caption entropy than 
				incorrect ones (\autoref{fig:Entropy} right), suggesting that broader 
				semantic coverage induced by hallucinations is positively associated with 
				successful reasoning. Rather than mere lexical variety, this diversity 
				reflects expanded semantic grounding that supports downstream task 
				performance. Further implementation details of similarity computation, entropy estimation, and statistical testing are provided in Appendix~\red{S8}. We give the following observations:
				
				\textbf{\ding{186} Hallucinations reshape semantic inputs.} 
				From~\autoref{fig:distribution}, we observe that hallucinated captions differ systematically from faithful ones in both mean similarity and distributional spread. 
				Hallucinated inputs exhibit wider variance and heavier tails in the embedding space, a difference that is statistically significant under paired t-tests ($p<0.01$). 
				This confirms that they introduce genuine input-level shifts rather than acting as redundant noise, providing models with additional anchors to explore alternative reasoning paths. 
				Importantly, this shift is consistently observed across multiple datasets, underscoring that input-level semantic reshaping is a general property of hallucinations rather than a dataset-specific artifact, and holds robustly across diverse modalities and domains.
				\begin{figure}[t]
					\centering
					\includegraphics[width=1.0\linewidth]{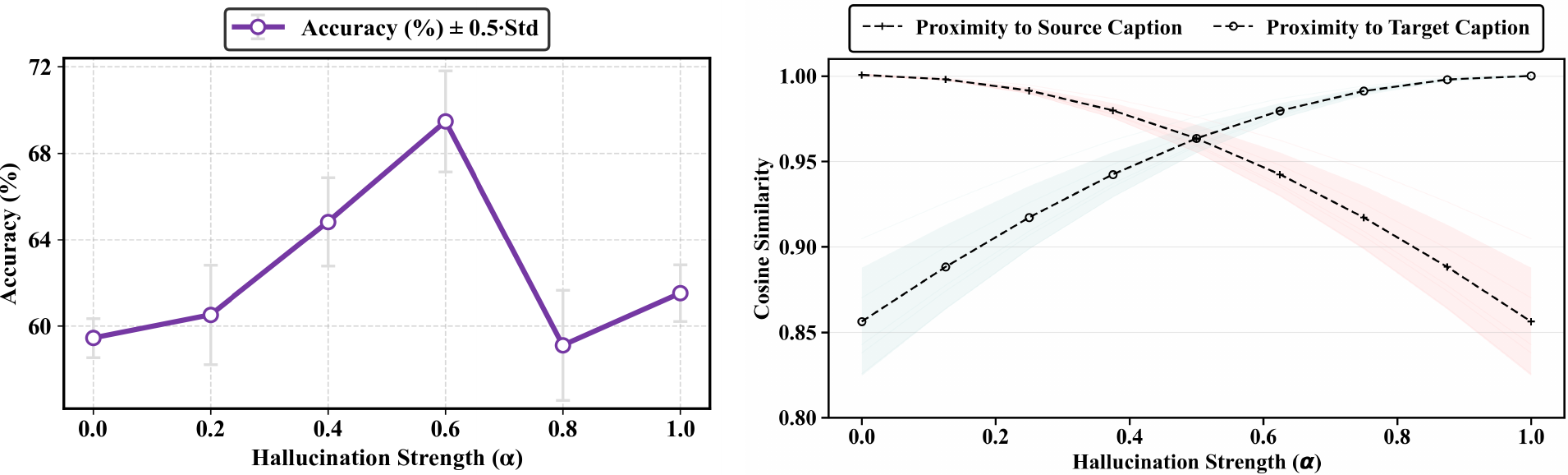}
					\caption{\textbf{Downstream task accuracy as a function of hallucination strength.}
						We gradually interpolate between faithful and hallucinated captions and evaluate downstream performance. 
						The results exhibit an inverted-U pattern: introducing moderate hallucination improves accuracy, 
						while excessive hallucination reduces it.}
					\label{fig:hallucination-strength}
					%\vspace{-7mm} 
				\end{figure}
				
				\textbf{\ding{187} \textbf{Hallucinations modulate reasoning dynamics.}} 
				As shown in \autoref{fig:Entropy} (left), hallucinated prompts alter the entropy of reasoning trajectories in a task-dependent manner. 
				On reasoning-heavy tasks such as BBBP, AntiCP2, and ProofWriter, hallucinations \emph{reduce} movement entropy, suggesting that they promote more convergent and stable inference. 
				Conversely, on structurally open-ended tasks such as SARA v3, hallucinations \emph{increase} entropy, enabling the model to explore a broader range of reasoning paths. 
				These differences are statistically significant (paired $t$-tests, $p<0.05$), indicating that hallucinations do not merely inject noise but actively reshape inference dynamics in ways that can either encourage convergence or support exploration, depending on the task.
				
				%	\begin{figure}[htbp]
					%	\centering
					%	\includegraphics[width=0.8\textwidth]{figures/figure2}
					%	\caption{\textbf{Distribution of caption embeddings.} 
						%		Hallucinated inputs exhibit wider semantic spread and longer tails. Stars indicate $p < 0.01$.}
					%	\label{fig:distribution}
					%	\vspace{-15pt}
					%	\end{figure}

				\textbf{\ding{188} \textbf{Correct predictions align with higher caption entropy.}} 
				As shown in \autoref{fig:Entropy} (right), hallucinated captions that lead to correct predictions consistently exhibit higher semantic entropy than those leading to incorrect predictions, across datasets such as GQA, Dex-Net, ISIC, and PlantVillage. 
				These differences are statistically significant ($p<0.05$), and the effect holds consistently across all evaluated datasets, underscoring that higher semantic diversity is a general marker of successful reasoning rather than a dataset-specific artifact. 
				Rather than mere lexical variety, this result highlights that semantic diversity in the latent space is a useful signal that supports accurate task performance.
				\begin{tcolorbox}
					\textbf{Takeaway \ding{183}.} Hallucinations consistently reshape inputs, modulate reasoning trajectories, and higher semantic diversity correlates with correct outcomes, indicating that their utility arises from broadening the semantic space rather than adding redundant noise.
				\end{tcolorbox}
				
				We further analyze hallucinations that arise within the reasoning chain itself. 
				By detecting the position of the first hallucinated token (early, middle, late), we observe consistent positional patterns: 
				early hallucinations tend to correlate with stronger downstream improvements. 
				This suggests that speculative cues introduced early in the reasoning process may shape the evolving inference trajectory. Detailed experimental results are provided in Appendix~\red{S4}.

				Intuitively, hallucinated captions expand the semantic hypothesis
				space available to the model. Under partial observability,
				faithful captions often describe only visible attributes,
				which may provide insufficient cues for reasoning.
				Hallucinated cues introduce speculative but task-relevant
				semantic anchors that guide the model toward plausible
				interpretations of the input. When these anchors align with
				the latent structure of the task, they help the model explore
				useful reasoning trajectories rather than remaining confined
				to incomplete observations.
				%	\begin{figure}[htbp]
					%		\centering
					%		\includegraphics[width=0.8\linewidth]{figures/figure6.pdf}
					%		\caption{\textbf{Downstream task accuracy as a function of hallucination strength.}
						%			We gradually interpolate between faithful and hallucinated captions and evaluate downstream performance. 
						%			The results exhibit an inverted-U pattern: introducing moderate hallucination improves accuracy, 
						%			while excessive hallucination reduces it. This confirms that moderate hallucination can serve as a beneficial signal, 
						%			whereas overly strong hallucination becomes detrimental.}
					%		\label{fig:hallucination-strength}
					%	\end{figure}

				\vspace{-4mm}
				\subsection{Case Study}
\begin{figure}[t]
					\centering
					\includegraphics[width=1.0\linewidth]{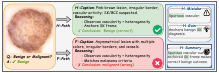}
					\caption{\textbf{Case study on \textsc{ISIC}.} 
						A hallucinated (H) caption introduces a spurious vascular cue that anchors the reasoning toward a seborrheic keratosis (SK) frame, ultimately yielding the correct benign diagnosis. In contrast, the faithful (F) caption confines reasoning to superficial features, leading to a malignant misclassification, highlighting hallucination’s potential as constructive guidance.}
					\label{fig:CaseStudy_ISIC}
					%\vspace{-6mm}
				\end{figure}

				\vspace{-2mm}
				\hspace*{2em}
				Beyond aggregate results, we present a case study to illustrate how 
				hallucinated captions reshape reasoning chains in practice. We select a 
				sample from the \textsc{ISIC} dataset, where the task is to determine 
				whether a skin lesion is benign or malignant. Given the same input, the 
				faithful (F) caption focuses on superficial attributes (e.g., asymmetry, 
				irregular borders), constraining the reasoning to melanoma-like criteria and 
				resulting in an incorrect malignant prediction. In contrast, the hallucinated (H) caption introduces a vascular cue suggestive 
				of a seborrheic keratosis (SK) frame. Although this cue is not strictly 
				faithful to the image, it provides a task-aligned semantic trigger that 
				reshapes the reasoning trajectory: intermediate steps increasingly anchor the 
				SK frame, ultimately leading to the correct benign diagnosis. This example demonstrates how hallucinations can supply additional semantic 
				anchors that steer the reasoning process toward more effective diagnostic 
				paths, rather than merely injecting noise. As illustrated in 
				\autoref{fig:CaseStudy_ISIC}, hallucinated captions can introduce 
				auxiliary but task-relevant cues that guide the reasoning chain toward the 
				correct outcome. More case studies across additional datasets are provided in Appendix~\red{S5}.
				
				\vspace{-4mm}
				\subsection{Ablation Study}
				\vspace{-2mm}
				\label{ablstudy}
				\hspace*{2em}
				To identify the conditions under which hallucinations provide the strongest utility, we conduct a set of ablation studies examining three key generation factors: sampling temperature, token budget, and hallucination intensity. These analyses reveal how different generation configurations shape the semantic expansion introduced by hallucinated captions and how such expansion translates into downstream performance gains.
				
				\textbf{Temperature.}
				We analyze how sampling temperature influences the usefulness of hallucinated captions.  
				As shown in \autoref{tab:temp_token}, all four datasets peak at $T=0.6$, 
				yielding the strongest and most consistent gains 
				(e.g., +11.76\% on \textsc{ISIC}, +14.68\% on \textsc{PlantVillage}, 
				+2.51\% on \textsc{Dex-Net}, +3.76\% on \textsc{AntiCP2}). 
				Lower temperatures ($T=0.0/0.3$) generate conservative captions with limited semantic expansion 
				(e.g., $-4.27$\% on \textsc{AntiCP2}, $-5.05$\% on \textsc{ISIC}), 
				while higher temperature ($T=0.9$) increases variance and reduces controllability 
				(e.g., $-5.00$\% on \textsc{AntiCP2}). 
				These observations indicate that a moderate temperature around $T=0.6$ achieves the most effective balance, 
				providing useful semantic enrichment without introducing excessive variability.

				\begin{table*}[htbp]
					\vspace{-10pt}
					\centering
					\scriptsize % 控制字体大小
					\setlength{\tabcolsep}{6pt}
					\renewcommand{\arraystretch}{1.15}
					\caption{\textbf{Hallucination-induced gain ($\Delta$) across temperature and token conditions.} 
						We report relative gain $\Delta=H-F$, isolating hallucination effects by eliminating baseline accuracy differences across datasets. \textbf{Bold} values mark the strongest gain and \underline{underlined} values the second-best.
					}
					\begin{tabular}{@{}lcccc|cccc@{}}
						\toprule
						\rowcolor{HeadGray}
						\textbf{Dataset} & \multicolumn{4}{c}{Temperature ($T$)} & \multicolumn{4}{c}{Token Length} \\
						\cmidrule(lr){2-5}\cmidrule(l){6-9}
						\rowcolor{HeadGray}
						& 0.0 & 0.3 & 0.6 & 0.9 & 128 & 256 & 512 & 1024 \\
						\midrule
						AntiCP2
						& -4.27 & \underline{+0.10} & \textbf{+3.76} & -5.00
						& +0.15 & \underline{+3.76} & \textbf{+4.48} & -0.14 \\
						PlantVillage
						& \underline{+2.30} & -4.46 & \textbf{+14.68} & +1.51
						& -4.66 & \textbf{+14.68} & +7.86 & \underline{+9.49} \\
						Dex-Net
						& +0.07 & \underline{+1.88} & \textbf{+2.51} & +0.14
						& +1.56 & \textbf{+2.51} & -2.04 & \underline{+1.93} \\
						ISIC
						& \underline{+9.26} & -5.05 & \textbf{+11.76} & +3.70
						& -2.10 & \textbf{+11.76} & \underline{+1.33} & -0.48 \\
						\bottomrule
					\end{tabular}
					\label{tab:temp_token}
					\vspace{-10pt} 
				\end{table*}

				\textbf{Maximum token budget.}
				We examine the impact of token length using temperature $T=0.6$ (\autoref{tab:temp_token}). 
				Short generations (128 tokens) limit semantic coverage and result in weak or inconsistent gains.  
				A budget of 256 tokens produces strong and stable improvements across all datasets.  
				Larger budgets (512 or 1024 tokens) can yield higher peaks but also introduce greater variance.  
				Ablating key hallucinated tokens results in a substantial accuracy drop, confirming the benefits of hallucinations rely on 
				sufficient and well-structured semantic cues.
				
				\textbf{Hallucination intensity.}
				To assess how hallucination strength affects downstream performance, 
				we generate captions with different hallucination levels using GPT-4o and group them into strong and weak categories.  
				We interpolate between faithful and hallucinated captions, and between strong and weak hallucinations, 
				then re-project them into SBERT space for alignment.  
				\autoref{fig:hallucination-strength} shows smooth semantic transitions (right) 
				and an inverted U-shaped pattern (left), where moderate hallucination intensity provides the most reliable accuracy improvements, while excessive intensity reduces controllability.  
				\begin{tcolorbox}
					\textbf{Takeaway \ding{184}.} Moderate hallucination levels provide the strongest gains, indicating that the benefits of hallucinated cues depend on controlled semantic expansion rather than excessive or weak generation.
				\end{tcolorbox}
				
				%	\begin{figure}[htbp]
					%		\centering
					%		\includegraphics[width=0.8\linewidth]{figures/figure6.pdf}
					%		\caption{\textbf{Downstream task accuracy as a function of hallucination strength.}
						%			We gradually interpolate between faithful and hallucinated captions and evaluate downstream performance. 
						%			The results exhibit an inverted-U pattern: introducing moderate hallucination improves accuracy, 
						%			while excessive hallucination reduces it. This confirms that moderate hallucination can serve as a beneficial signal, 
						%			whereas overly strong hallucination becomes detrimental.}
					%		\label{fig:hallucination-strength}
					%	\end{figure}
				
				%\begin{figure}[t]
				%	\centering
				%	\includegraphics[width=1.0\linewidth]{figures/figure6.pdf}
				%	\caption{\textbf{Downstream task accuracy as a function of hallucination strength.}
					%		We gradually interpolate between faithful and hallucinated captions and evaluate downstream performance. 
					%		The results exhibit an inverted-U pattern: introducing moderate hallucination improves accuracy, 
					%		while excessive hallucination reduces it. This confirms that moderate hallucination can serve as a beneficial signal, 
					%		whereas overly strong hallucination becomes detrimental.}
				%	\label{fig:hallucination-strength}
				%	\vspace{-6mm}
				%\end{figure}

				\vspace{-4mm}
				\section{Discussion and Conclusion} \label{Discussion_and_Conclusion}
				\vspace{-4mm}
				\hspace*{2em}
				Our findings point to a dual characterization of post-hallucination reasoning. 
				The \emph{faithful path} encourages exploitation, grounding the model in 
				verified evidence and producing precise but narrow predictions. The 
				\emph{hallucinated path} promotes exploration, expanding the hypothesis 
				space through speculative but task-relevant cues that occasionally unlock 
				shortcuts unavailable to faithful inputs. Hallucinations are thus not merely 
				errors but alternative signals that broaden the model's inference landscape. 
				Their benefits, however, depend critically on control: moderate hallucination 
				intensity enriches semantics without destabilizing inference, whereas excessive 
				or misaligned hallucinations degrade reliability. Temperature, token budget, 
				and interpolation strength provide practical levers for tuning this 
				balance across datasets, models, and decoding settings, and fallback mechanisms 
				to the faithful path help manage risk.
				
				\textbf{Limitations.}
				Although our study reveals consistent patterns of PHR across tasks and models, several limitations remain. 
				First, our evaluation is conducted on a fixed set of benchmarks and may not fully capture the diversity of real-world multimodal reasoning scenarios. 
				Second, our interventions operate primarily at the caption level, whereas hallucinations may also arise within intermediate reasoning steps or latent representations. 
				Third, while hallucinated cues can sometimes assist reasoning under partial observability, they may also introduce spurious signals in other settings. 
				Future work should investigate broader task domains and develop principled mechanisms to better control hallucination induced semantic expansion in multimodal reasoning systems.
				
				\textbf{Broader Impact.}
				This work contributes to a deeper understanding of hallucination behavior in 
				multimodal models by examining how hallucinated semantics interact with 
				downstream reasoning. Importantly, our findings should not be interpreted as 
				advocating hallucination as a deliberate inference strategy. Rather, PHR 
				appears to arise as a by-product of reasoning under incomplete evidence. Understanding this phenomenon may inform future methods that 
				balance exploratory semantic cues with reliable grounding and verification.
				
				We hope this work encourages the community to further investigate the role of hallucinated semantics in model reasoning and develop principled methods to understand and control PHR.
				
				\vspace{-3mm}
				\section*{Acknowledgements}
				\hspace*{2em}
				We thank the reviewers and area chair for their constructive feedback. 
				%We also thank our colleagues for helpful discussions and suggestions.

					\newpage

\bibliographystyle{splncs04}
\bibliography{refs}

	\clearpage

	\appendix
	
	    \renewcommand{\thesection}{S\arabic{section}}
	\renewcommand{\thetable}{S\arabic{table}}
	\renewcommand{\thefigure}{S\arabic{figure}}
	\setcounter{table}{0}
	\setcounter{figure}{0}
	\begin{center}
		\large\textbf{SUMMARY OF THE APPENDIX}
	\end{center}
	This supplementary contains additional details for ECCV 2026, titled \textit{``HIVE: Understanding Post Hallucination Reasoning in Vision Language Models''}. The supplementary is organized as follows:
	\begin{itemize}
		\item \S\ref{app:statistics} reports the \textbf{significance and robustness analysis}. It includes mean$\pm$std, $\Delta$(H--F), and $p$-values across datasets.
		\item \S\ref{app:Prompt_Templates} lists the \textbf{prompt templates}. Each dataset has a role prompt, a generation prompt, and an evaluation prompt.
		\item \S\ref{app:scaling} studies the \textbf{effect of model scale}, showing that hallucination gains are non-monotonic across Qwen2.5-VL sizes.
%		\item \S\ref{sec:rand-checker} presents the \textbf{random checker control}. It confirms that improvements are not due to arbitrary filtering.
%		\item \S\ref{sec:token-ablation} reports the \textbf{token ablation study}. Core hallucinated tokens are shown to be necessary for success.
		\item \S\ref{app:positional} analyzes the \textbf{positional effects of hallucinations in reasoning chains}. It studies how the position of hallucinated tokens (early, middle, late) influences downstream task performance across datasets and models.
		\item \S\ref{app:case} presents the \textbf{case studies}. It provides qualitative examples on DexNet, BBBP, and PlantVillage, showing how hallucinated captions act as anchors that guide reasoning toward correct outcomes
		\item \S\ref{sec:exp-setup} summarizes the \textbf{datasets, models, and evaluation protocols}.
		\item \S\ref{sec:icd} explains the \textbf{caption discriminator}. Three complementary factuality verifiers are described in detail.
		\item \S\ref{app:analysis_setup} provides the \textbf{analysis setup}. Input-level, process-level, and output-level analysis pipelines are described separately.
		\item \S\ref{app:analysis_convergence} provides the \textbf{convergence and similarity analysis}. Both intra-chain and inter-chain convergence are reported.
		\item \S\ref{sec:additional_robustness} reports the \textbf{additional robustness experiments} on MMStar and MMBench, showing consistent gains across perception and reasoning subsets.
	\item \S\ref{Appendix:License} lists the \textbf{dataset and model licenses}.
	\item \S\ref{appendix:disclosure} states that \textbf{GPT-5 was used only for grammar checking}.
	\end{itemize}

	\section{Significance and Robustness}
	\label{app:statistics}

	% \begin{table}[htbp]
		% 	\centering
		% 					\caption{\textbf{Statistical significance of hallucination-induced gains.} 
			% 		We report mean$\pm$std over 5 runs. $\Delta$(H–F) denotes accuracy gain. 
			% 		$p$-values are from two-sided paired $t$-tests; significant results ($p<0.05$) are bolded.}
		% 	\resizebox{\textwidth}{!}{
			% 		\begin{tabular}{lcccccc}
				% 			\toprule
				% 			Dataset & Domain & Faithful (F) & Hallucinated (H) & $\Delta$(H–F) & $p$ \\
				% 			\midrule
				% 			AntiCP2       & Protein        & 0.5459$\pm$0.0067 & 0.5835$\pm$0.0126 & +0.0376 & \textbf{0.00036} \\
				% 			BBBP          & Drug property  & 0.6167$\pm$0.0264 & 0.6833$\pm$0.0118 & +0.0667 & \textbf{0.00481} \\
				% 			CodeXGLUE     & C++ code       & 0.5515$\pm$0.0326 & 0.5275$\pm$0.0227 & –0.0240 & 0.102 \\
				% 			SARA\_V3      & Law reasoning  & 0.6293$\pm$0.0084 & 0.6224$\pm$0.0161 & –0.0069 & 0.147 \\
				% 			ProofWriter   & Logic          & 0.6949$\pm$0.0125 & 0.7500$\pm$0.0217 & +0.0551 & \textbf{1.57E-05} \\
				% 			GQA           & VQA multimodal & 0.7136$\pm$0.0040 & 0.7422$\pm$0.0078 & +0.0286 & \textbf{0.00088} \\
				% 			DexNet        & Robotics       & 0.5325$\pm$0.0045 & 0.5576$\pm$0.0116 & +0.0251 & \textbf{0.00038} \\
				% 			ISIC          & Medical        & 0.6388$\pm$0.0126 & 0.7564$\pm$0.0248 & +0.1176 & \textbf{0.00151} \\
				% 			PlantVillage  & Agriculture    & 0.6273$\pm$0.0249 & 0.7741$\pm$0.0211 & +0.1468 & \textbf{3.42E-09} \\
				% 			\bottomrule
				% 	\end{tabular}}
		% 	\label{full_table}
		% \end{table}
	
	\begin{table}[htbp]
		\centering
		\caption{\textbf{Statistical significance of hallucination-induced gains.} Mean$\pm$Std over 5 runs. $\Delta$(H–F) denotes accuracy gain. Two-sided paired $t$-test $p$-values; significant results ($p{<}0.05$) are bolded.}
		\label{full_table}
		\resizebox{0.95\textwidth}{!}{%
			{\renewcommand{\arraystretch}{1.3}
				\begin{tabular}{lccccc}
					\hline
					\rowcolor{CadetBlue!20}
					\textbf{Dataset} & \textbf{Domain} & \textbf{Faithful (F)} & \textbf{Hallucinated (H)} & \boldmath$\Delta$(H–F)\unboldmath & \textbf{$p$} \\
					\hline
					\rowcolor{gray!10}
					AntiCP2      & Protein        & $0.5459{\pm}0.0067$ & $0.5835{\pm}0.0126$ & $+0.0376$ & \textbf{0.00036} \\
					BBBP         & Drug property  & $0.6167{\pm}0.0264$ & $0.6833{\pm}0.0118$ & $+0.0667$ & \textbf{0.00481} \\
					\rowcolor{gray!10}
					CodeXGLUE    & C++ code       & $0.5515{\pm}0.0326$ & $0.5275{\pm}0.0227$ & $-0.0240$ & $0.102$ \\
					SARA\_V3     & Law reasoning  & $0.6293{\pm}0.0084$ & $0.6224{\pm}0.0161$ & $-0.0069$ & $0.147$ \\
					\rowcolor{gray!10}
					ProofWriter  & Logic          & $0.6949{\pm}0.0125$ & $0.7500{\pm}0.0217$ & $+0.0551$ & \textbf{$1.57{\times}10^{-5}$} \\
					GQA          & VQA multimodal & $0.7136{\pm}0.0040$ & $0.7422{\pm}0.0078$ & $+0.0286$ & \textbf{0.00088} \\
					\rowcolor{gray!10}
					DexNet       & Robotics       & $0.5325{\pm}0.0045$ & $0.5576{\pm}0.0116$ & $+0.0251$ & \textbf{0.00038} \\
					ISIC         & Medical        & $0.6388{\pm}0.0126$ & $0.7564{\pm}0.0248$ & $+0.1176$ & \textbf{0.00151} \\
					\rowcolor{gray!10}
					PlantVillage & Agriculture    & $0.6273{\pm}0.0249$ & $0.7741{\pm}0.0211$ & $+0.1468$ & \textbf{$3.42{\times}10^{-9}$} \\
					\hline
			\end{tabular}}
		}
	\end{table}

	\hspace*{2em}
	All reported results are averaged over five independent runs with different random seeds
	and presented as mean$\pm$std. We conduct two-sided paired $t$-tests to compare faithful (F)
	and hallucinated (H) inputs. \autoref{full_table} reports the full results across all nine datasets, including
	mean$\pm$std, relative gain $\Delta$(H--F), and $p$-values. The majority of tasks exhibit statistically
	significant gains ($p{<}0.05$ or $p{<}0.01$). For example, hallucinations yield large and consistent
	improvements on perception-heavy datasets such as ISIC ($+11.8\%$, $p=0.0015$) and
	PlantVillage ($+14.7\%$, $p{<}10^{-8}$), while rule-driven tasks such as CodeXGLUE and SARA
	show negligible or non-significant differences ($p{>}0.1$). These results confirm that the reported
	improvements are statistically reliable rather than random variation.
	
	\section{Prompt Templates}
		\begin{table}[!htbp]
		\centering
		\renewcommand{\arraystretch}{1.15}
		\scriptsize
		\caption{\textbf{Prompts used across datasets.} Each dataset is paired with a role prompt, a generation prompt, and an evaluation prompt, ensuring task-specific context and consistency.}
		\begin{tabular}{p{0.11\textwidth}p{0.20\textwidth}p{0.27\textwidth}p{0.35\textwidth}}
			\toprule
			\textbf{Dataset} & \textbf{Role prompt} & \textbf{Generation prompt} & \textbf{Evaluation prompt} \\
			\midrule
			AntiCP2 & You are a protein science expert. &
			\texttt{\{Sign\}\textbackslash n} Describe this in natural language: &
			You are a peptide bioinformatics expert responsible for evaluating short peptide sequences for the presence or absence of anticancer activity. Answer: yes or no. Then provide a step-by-step reasoning process. \\
			\addlinespace[1pt]
			BBBP & You are an expert in drug discovery. &
			\texttt{\{Sign\}\textbackslash n} Describe this in natural language: &
			Does the molecule have the ability to penetrate the blood-brain barrier? Answer: yes or no. Then provide a step-by-step reasoning process. \\
			\addlinespace[1pt]
			CodeX. & You are a software security expert and professor. &
			\texttt{\{Sign\}\textbackslash n} Describe this in natural language: &
			You are a software security expert and professor. Does the following C function contain a security vulnerability? Answer: yes or no. Then provide a step-by-step reasoning process. \\
			\addlinespace[1pt]
			SARA\_V3 & You are a legal expert. &
			\texttt{\{Sign\}\textbackslash n} Describe this in natural language: &
			You are a legal reasoning assistant. Determine whether the following legal claim is supported by the facts. Answer: yes or no. Then provide a step-by-step reasoning process. \\
			\addlinespace[1pt]
			Proof & You are an assistant for reasoning. &
			\texttt{\{Sign\}\textbackslash n} Describe this in natural language: &
			You are a logical reasoning assistant. Determine whether the statement is entailed by the given context of facts and rules. Answer: yes or no. Then provide a step-by-step reasoning process. \\
			\addlinespace[1pt]
			GQA & You are a reason expert. &
			Describe this image in natural language: &
			You are a visual reasoning expert. Answer the question based on the image. Answer: yes or no. Then provide a step-by-step reasoning process. \\
			\addlinespace[1pt]
			Dex\_Net & You are an expert in robotic grasp assessment. &
			Describe this image in natural language: &
			You are a senior robotic manipulation engineer specializing in parallel-jaw grasp planning. Answer: yes or no. Then provide a step-by-step reasoning process. \\
			\addlinespace[1pt]
			ISIC & You are an expert dermatoscopist. &
			Describe this image in natural language: &
			You are an expert dermatoscopist. Based on this image, decide whether the lesion is malignant (melanoma) or benign. Answer: yes (malignant melanoma) or no (benign). Then provide a step-by-step reasoning process. \\
			\addlinespace[1pt]
			PlantV. & You are a seasoned plant pathologist for solanaceous crops &
			Describe this image in natural language: &
			You are an expert plant pathologist who diagnoses tomato foliar diseases. Decide whether it shows early blight or late blight: reply yes if it is early blight and no if it is late blight. Answer: yes or no. Then provide a step-by-step reasoning process. \\
			\bottomrule
		\end{tabular}
		\setlength{\tabcolsep}{4pt}
		\label{tab:prompt-templates}
	\end{table}
	\hspace*{2em}
	To ensure consistency across benchmarks, we design unified prompt templates that
	follow a three-part structure: a \emph{role prompt}, a \emph{generation prompt},
	and an \emph{evaluation prompt}. The role prompt assigns the model an expert
	identity tailored to the domain (e.g., drug discovery, legal reasoning, medical
	diagnosis). The generation prompt asks the model to verbalize the raw input
	(\{Sign\}) into natural language, thereby producing either a faithful or a
	hallucinated caption. Finally, the evaluation prompt specifies the downstream
	task, which always requires a binary decision (yes/no) together with a
	step-by-step reasoning chain. This design ensures that the only experimental
	variable is the type of caption (faithful vs.\ hallucinated), while all other
	aspects of the prompt remain controlled. 
	
	\autoref{tab:prompt-templates} lists the complete templates used for all nine
	datasets. These include both text-based tasks (AntiCP2, BBBP, CodeXGLUE, SARA,
	ProofWriter) and multimodal tasks (GQA, DexNet, ISIC, PlantVillage). The
	templates were fixed across all models and experiments, so that observed
	differences can be attributed solely to the presence or absence of hallucinated
	semantics.
	\label{app:Prompt_Templates}

	\section{Effect of model scale}
	\label{app:scaling}
		\hspace*{2em}
We evaluate Qwen2.5-VL models at three different scales (7B, 32B, 72B).
As shown in \autoref{tab:qwen-scaling}, the impact of hallucinations is not monotonic with scale.
The 32B model benefits substantially (+9.4\%), while the 7B and 72B models show slight drops.
This suggests that scale alone does not determine hallucination effectiveness:
intermediate-scale models may gain from additional semantic cues,
whereas very large models may already saturate on faithful inputs,
making further hallucinations redundant or even distracting.

\section{Positional Effects of Hallucinations in Reasoning Chains}
\label{app:positional}
%\begin{table}[t]
%	\centering
%	\caption{Positional effects of hallucinations within reasoning chains.
%		Accuracy changes ($\Delta$) are measured relative to the
%		\textit{none} condition.}
%	\label{tab:positional_effects}
%	\resizebox{0.82\textwidth}{!}{%
%		{\renewcommand{\arraystretch}{1.3}
%			\begin{tabular}{lcccc}
%				\hline
%				\rowcolor{CadetBlue!20}
%				\textbf{Dataset / Model}
%				& \textbf{None}
%				& \textbf{Early} \boldmath$\Delta$\unboldmath
%				& \textbf{Late} \boldmath$\Delta$\unboldmath
%				& \textbf{Middle} \boldmath$\Delta$\unboldmath \\
%				\hline
%				
%				\rowcolor{gray!10}
%				BBBP + GPT-4o
%				& $65.22$
%				& $\mathbf{+23.11}$
%				& $+8.23$
%				& $-15.22$ \\
%				
%				BBBP + Claude-3
%				& $52.93$
%				& $+21.42$
%				& $+43.07$
%				& $\mathbf{+44.85}$ \\
%				
%				\rowcolor{gray!10}
%				ISIC + GPT-4o
%				& $78.47$
%				& $\mathbf{+13.75}$
%				& $+5.31$
%				& $+4.20$ \\
%				
%				ISIC + Claude-3
%				& $80.05$
%				& $+3.96$
%				& $\mathbf{+16.38}$
%				& $+10.12$ \\
%				\hline
%			\end{tabular}
%		}
%	}
%\end{table}

	\begin{table}[t]
		\centering
		\caption{\textbf{Scaling results within the Qwen2.5-VL family.}
			Hallucination effects are non-monotonic: smaller and medium-large models benefit,
			while very large models show volatility or saturation.}
		\label{tab:qwen-scaling}
		
		\small
		\resizebox{0.82\textwidth}{!}{%
			\begin{tabular}{lccc}
				\hline
				\rowcolor{headergray}
				\textbf{Model} & \textbf{Faithful (F)} & \textbf{Hallucinated (H)} & $\boldsymbol{\Delta}(\mathrm{H}{-}\mathrm{F})$ \\
				\hline
				\rowcolor{gray!10}
				Qwen2.5-VL-7B
				& $0.5898{\pm}0.0000$
				& $0.5276{\pm}0.0000$
				& $-0.0622$ \\
				
				Qwen2.5-VL-32B
				& $0.5935{\pm}0.0068$
				& $\mathbf{0.6874{\pm}0.0409}$
				& $\mathbf{+0.0939}$ \\
				
				\rowcolor{gray!10}
				Qwen2.5-VL-72B
				& $\mathbf{0.7349{\pm}0.0171}$
				& $0.6856{\pm}0.0167$
				& $-0.0493$ \\
				\hline
			\end{tabular}%
		}
	\end{table}

	\begin{table}[t]
		\centering
		\caption{\textbf{Positional effects of hallucinations within reasoning chains.}
			Accuracy changes ($\Delta$) are measured relative to the
			\textit{none} condition.}
		\label{tab:positional_effects}
		
		\small
		\resizebox{0.82\textwidth}{!}{%
			\begin{tabular}{lcccc}
				\hline
				\rowcolor{headergray}
				\textbf{Dataset / Model} & \textbf{None} & \textbf{Early $\boldsymbol{\Delta}$} & \textbf{Late $\boldsymbol{\Delta}$} & \textbf{Middle $\boldsymbol{\Delta}$} \\
				\hline
				\rowcolor{gray!10}
				BBBP + GPT-4o
				& $65.22$
				& $\mathbf{+23.11}$
				& $+8.23$
				& $-15.22$ \\
				
				BBBP + Claude-3
				& $52.93$
				& $+21.42$
				& $+43.07$
				& $\mathbf{+44.85}$ \\
				
				\rowcolor{gray!10}
				ISIC + GPT-4o
				& $78.47$
				& $\mathbf{+13.75}$
				& $+5.31$
				& $+4.20$ \\
				
				ISIC + Claude-3
				& $80.05$
				& $+3.96$
				& $\mathbf{+16.38}$
				& $+10.12$ \\
				\hline
			\end{tabular}%
		}
	\end{table}

	\hspace*{2em}
While the main paper focuses on hallucinations introduced through caption-based interventions, hallucinations may also emerge within the model’s reasoning chain itself. To better understand their potential impact on downstream predictions, we conduct an additional analysis examining the positional effects of hallucinated tokens within generated reasoning trajectories. 

Specifically, we detect hallucinated tokens in the reasoning chains produced by the models and group samples according to the position of the first hallucination in the sequence, categorized as \textit{early}, \textit{middle}, \textit{late}, or \textit{none}. For each category, we measure the resulting task accuracy and compare it with the baseline condition where no hallucination occurs. Experiments are conducted on two representative datasets (BBBP and ISIC) using two models (GPT-4o and Claude-3 Sonnet). The results are summarized in Table~\ref{tab:positional_effects}, which reports the accuracy differences relative to the \textit{none} condition.

		\section{Qualitative Case Studies}
		\label{app:case}
\begin{figure}[t]
			\centering
			\includegraphics[width=1.0\linewidth]{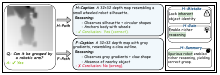}
			\caption{\textbf{Case study on DexNet.} 
				A hallucinated (H) caption misinterprets the depth map as a robot silhouette with wheels, anchoring reasoning toward a graspable object and yielding the correct answer.
				In contrast, the faithful (F) caption only notes gray gradients and claw-like shapes, failing to establish object identity and leading to an incorrect ``No'' prediction.
				This example shows how hallucinated cues, though factually incorrect, can enrich reasoning and enable correct decisions.}
			\label{fig:CaseStudy_DexNet}
			\vspace{-10pt} 
		\end{figure}

		\textbf{DexNet: Robotic Grasping.} \autoref{fig:CaseStudy_DexNet} shows a robotic grasping case from DexNet. 
		The hallucinated (H) caption mistakenly interprets the depth map as a wheeled robot silhouette, 
		but this spurious cue provides a concrete object anchor that enables correct reasoning for graspability. 
		In contrast, the faithful (F) caption only describes gray gradients and claw-like shapes, 
		failing to establish object identity and thus leading to the wrong ``No'' prediction. 
		This case illustrates how hallucinations, though factually incorrect, can enrich the reasoning space and support the correct decision.
		
\begin{figure}[t]
			\centering
			\includegraphics[width=1.0\linewidth]{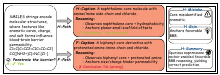}
			\caption{\textbf{Case study on BBBP.} 
				A hallucinated (H) caption incorrectly identifies the molecule as naphthalene-based but introduces a hydrophobic anchor favoring blood–brain barrier permeability, leading to the correct ``Yes'' outcome. In contrast, the faithful (F) caption focuses on a biphenyl scaffold with protonated amine, anchoring reasoning on size and charge constraints and resulting in the wrong ``No'' prediction.
				This example illustrates how aromatic cues, though factually mistaken, can provide constructive anchors that guide reasoning toward correct molecular permeability.}
			\label{fig:CaseStudy_BBBP}
			\vspace{-10pt} 
		\end{figure}

		\textbf{BBBP: Molecular Permeability.} \autoref{fig:CaseStudy_BBBP} presents a molecular classification example from BBBP. 
		The hallucinated (H) caption misidentifies the scaffold as naphthalene, 
		but this cue anchors reasoning toward favorable hydrophobicity, 
		guiding the model to correctly predict blood–brain barrier penetration. 
		Meanwhile, the faithful (F) caption emphasizes a biphenyl scaffold with protonated amine, 
		anchoring reasoning on size/charge constraints and resulting in the wrong ``No'' prediction. 
		This case demonstrates that even erroneous aromatic anchors can serve as constructive signals for correct permeability classification.
		
\begin{figure}[t]
			\centering
			\includegraphics[width=1.0\linewidth]{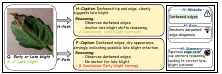}
			\caption{\textbf{Case study on PlantVillage.} 
				A hallucinated (H) caption highlights a spurious cue darkened tip and edges that anchors reasoning toward late blight, ultimately yielding the correct diagnosis. 
				In contrast, the faithful (F) caption notes the same darkened edges but lacks an explicit anchor for late blight, leading to an incorrect early-blight classification. 
				This example illustrates how hallucinations, even when grounded in partially misleading features, can provide decisive anchors that guide reasoning toward the correct outcome.}
			\label{fig:CaseStudy_PlantVillage}
			\vspace{-10pt} 
		\end{figure}

		\textbf{PlantVillage: Crop Disease Recognition.} \autoref{fig:CaseStudy_PlantVillage} illustrates a crop disease recognition task. 
		The hallucinated (H) caption highlights darkened tips and edges, anchoring reasoning toward late blight and producing the correct diagnosis. 
		In contrast, the faithful (F) caption notes similar edge darkening but does not explicitly anchor late blight, 
		leading to the incorrect early-blight decision. 
		This case underscores that hallucinations, even when based on spurious cues, can act as decisive anchors that steer reasoning toward the correct outcome.
		
		\textbf{Summary of Case Study.} Across DexNet, BBBP, and PlantVillage, a consistent pattern emerges: 
		hallucinated captions often introduce spurious or factually mistaken cues 
		(e.g., a robot silhouette, a naphthalene core, or darkened leaf edges). 
		Yet these cues act as decisive anchors that expand the reasoning space, 
		providing additional structure that guides the model toward the correct outcome. 
		By contrast, faithful captions though factually accurate may lack sufficient anchoring, 
		causing reasoning to remain shallow and sometimes incorrect. 
		These case studies highlight hallucination’s constructive potential: 
		even when imperfect, hallucinations can inject inductive signals that improve decision quality.
		
		\section{Experimental setup}
		\label{sec:exp-setup}
			\hspace*{2em}
		\textbf{Datasets.} 
		We conduct experiments on 9 datasets spanning both textual and multimodal domains. 
		\textit{Text (5):} 
		AntiCP2~\cite{AntiCP2} (antimicrobial peptide classification), 
		BBBP~\cite{BBBP} (blood–brain barrier penetration), 
		CodeXGLUE~\cite{CodeXGLUE} (C++ exception prediction), 
		SARA~\cite{SARA} (legal reasoning), 
		ProofWriter~\cite{ProofWriter} (logic-based natural language inference). 
		\textit{Multimodal (4):} 
		GQA~\cite{GQA} (visual question answering), 
		DexNet~\cite{Dex_Net} (depth-based robotic grasping), 
		ISIC~\cite{ISIC} (skin-lesion classification), 
		PlantVillage~\cite{PlantVillage} (plant-disease recognition from RGB images).
		
		\textbf{Models.} 
		We evaluate 9 large language models, covering both proprietary and open-source systems: 
		\textit{Closed-source:} GPT-4o, GPT-3.5-turbo, Claude 3 Sonnet, Gemini 2.0 Flash, O3. \\
		\textit{Open-source:} DeepSeek-V3, DeepSeek-R1, Mistral Large, Qwen-VL
		
		\textbf{Evaluation.} 
		For binary classification tasks, we report Accuracy as the primary metric, 
		ensuring consistency across datasets and model families.
		
		\textbf{Statistics.}
		Unless otherwise noted, we report mean$\pm$std over 5 independent runs.
		For each dataset, we conduct two-sided paired \emph{t}-tests to compare faithful vs.\ hallucinated inputs. 
		Statistical significance is reported at conventional thresholds ($p{<}0.01$); 
		full results and additional details are provided in Appendix~\ref{app:statistics}.
		Token budget is implemented as a maximum generation length, though generations may terminate earlier.
		
		\section{Implementation of Caption Discriminator}
		\label{sec:icd}
			\hspace*{2em}
		We implement three complementary modules to assess the factual plausibility of
		hallucinated captions. Each module is motivated by prior work on self-consistency,
		fine-grained fact-checking, and paraphrase-based semantic validation.
		
		\textbf{Fine-Grained Factuality Verifier.} Motivated by the fine-grained evaluation perspective of Factcheck-Bench
		\cite{check2}, this module decomposes each caption into individual factual
		claims using sentence segmentation. Each claim is independently verified by a
		large language model with a structured prompt that returns a binary verdict
		(True/False), a confidence score (0--1), and a short justification. The final score averages the confidence of verified True claims, with a penalty for detected False claims. This design enables auditing hallucinations at the individual claim level,
		rather than only at the aggregated whole-caption level used previously in prior evaluations.

		\textbf{Self-Evaluation Factuality Verifier.} Inspired by self-consistency approaches such as SelfCheckGPT
		\cite{manakul2023selfcheckgpt}, this module prompts the model to directly
		self-assess the factual correctness of an answer (with optional question
		context) in multiple practical scenarios. The model outputs a binary verdict with confidence and explanation.
		Compared to the fine-grained verifier, this method is lightweight and evaluates
		factuality at the whole-answer level. We also support a multimodal variant that
		incorporates image inputs when available across diverse evaluation settings.

		\textbf{Paraphrase-Consistency Verifier.} Following the idea of leveraging paraphrasing and question generation for
		semantic consistency~\cite{check3}, this module generates two paraphrases of
		the original caption while strictly preserving meaning. The paraphrases serve
		only as auxiliary evidence to clarify intent, while the factuality decision
		always prioritizes the original caption. A fact-adjudication prompt then
		produces a binary verdict, confidence, and concise reasoning. This consistency check reduces prompt variance and stabilizes factuality judgments overall accuracy.

		\textbf{Caption Discriminator.} Together, these three discriminators provide complementary perspectives:
		(I) fine-grained claim verification, (II) holistic self-evaluation, and
		(III) paraphrase-assisted consistency. In our experiments, we include a \emph{random checker} baseline (accept/reject uniformly at random) and ensemble variants combining multiple verifiers.

		\section{Implementation Details of Analysis Setup}
			\hspace*{2em}
		\label{app:analysis_setup}
		\textbf{Input-level.} To quantify the differences between faithful and hallucinated captions, and
		between correct and incorrect predictions, we adopt the following analysis
		pipeline. For each dataset, we collect hallucinated caption embeddings produced by the
		generation model. When multiple runs are available, we resolve embeddings by
		searching run-specific directories or aggregated  to ensure
		consistent coverage. Captions and embeddings are aligned with prediction labels,
		with samples truncated if necessary to guarantee matching length. High-dimensional embeddings are projected to a three-dimensional latent space
		using principal component analysis (PCA) with full SVD. This preserves the
		dominant semantic directions while removing redundant variance, facilitating
		density estimation. We estimate local distributional entropy of the embeddings by fitting a Gaussian
		kernel density estimator (KDE) with fixed bandwidth. For each sample, the
		negative log-density serves as its entropy value, reflecting whether it lies in
		a dense or sparse region of the semantic space. We report mean and standard
		deviation of entropy separately for correct and incorrect predictions.
		To assess the significance of differences between correct and incorrect groups,
		we conduct two-sided independent-sample \emph{t}-tests without assuming equal
		variance. We report the test statistic and $p$-value for each dataset. Results
		are aggregated across all nine benchmarks and summarized in
		Appendix~\S\ref{app:statistics}. This procedure provides a principled way to examine how hallucinations reshape
		semantic distributions at the input level, modulate reasoning trajectories, and
		correlate with prediction accuracy through entropy-based analysis.
		
		\textbf{Process-level.} To examine how hallucinations modulate inference dynamics, we quantify the entropy of reasoning-chain embeddings. 
		For each input, we record the hidden-state representations of step-wise reasoning trajectories under both faithful (F) and hallucinated (H) captions. We then project these embeddings into a lower-dimensional space via principal component analysis (PCA) and estimate their density distribution using kernel density estimation (KDE). 
		The negative log-likelihood of KDE outputs serves as an entropy measure, capturing the dispersion of reasoning movements across steps. 
%		For each dataset, we compute the mean entropy under F and H conditions, and report their differences as shown in \autoref{fig:Entropy} (left). 
		Paired two-sided $t$-tests are applied to assess statistical significance ($p<0.05$). 
		This measurement allows us to characterize whether hallucinations encourage more convergent reasoning trajectories (lower entropy) or diversify inference paths (higher entropy), depending on the task structure.
		
		\textbf{Output-level.} To analyze the semantic effect of hallucinations, we estimate the entropy of
		caption embeddings under the hallucinated (H) condition and compare between
		correct and incorrect predictions. 
		For each dataset, we collect the
		OpenCLIP embeddings of hallucinated captions ($C_H$). Predictions and gold
		labels are aligned with these embeddings by matching the number of instances. To improve stability and reduce noise
		in density estimation, embeddings are projected into a 3-dimensional latent
		space using Principal Component Analysis (PCA). This preserves the dominant
		variance directions while mitigating the curse of dimensionality. 
		We adopt Kernel Density Estimation (KDE)
		with a Gaussian kernel (bandwidth = 0.5) to approximate the underlying semantic
		distribution. For each sample, we compute the negative log-likelihood under the
		KDE as a proxy for semantic entropy. 
		We split samples into two
		groups based on prediction correctness and compute mean $\pm$ standard deviation
		of entropy for each group. Statistical differences are assessed using two-sided
		$t$-tests under unequal variance assumptions. 
		This procedure yields a robust measure of semantic diversity in hallucinated
		captions, allowing us to test whether correct predictions are associated with
		higher entropy than incorrect ones. \autoref{tab:entropy_analysis} summarizes caption entropy under hallucinated (H) inputs, 
		split by correct vs.\ incorrect predictions. 
		We find that correct predictions generally align with higher entropy, 
		with significant differences on four multimodal datasets (GQA, DexNet, ISIC, PlantVillage). 
		These results confirm that semantic diversity is a reliable marker of successful reasoning 
		rather than a dataset-specific artifact.
		
		% \begin{table}[t]
			% 	\centering
			% 	\caption{\textbf{Caption entropy analysis (H condition).} 
				% 		We compare entropy between correct and incorrect predictions. 
				% 		Values show mean entropy for each group, their difference, and statistical tests (two-sided $t$-test). 
				% 		Significant results ($p<0.05$) are bolded.}
			% 	\label{tab:entropy_analysis}
			% 	\begin{tabular}{lccccc}
				% 		\toprule
				% 		Dataset & Correct & Wrong & $\Delta$ (C--W) & $t$-stat & $p$-value \\
				% 		\midrule
				% 		AntiCP2      & 0.861 & 0.856 & +0.005  & 0.32  & 0.749 \\
				% 		BBBP         & 0.886 & 0.960 & $-$0.074 & -2.26 & \textbf{0.032} \\
				% 		CodeXGLUE    & 0.901 & 0.897 & +0.004  & 0.30  & 0.768 \\
				% 		SARA\_V3     & 1.030 & 1.004 & +0.025  & 1.91  & 0.060 \\
				% 		ProofWriter  & 0.975 & 0.998 & $-$0.023 & -1.14 & 0.262 \\
				% 		GQA          & 1.150 & 1.048 & +0.102  & 4.61  & \textbf{1.4e-5} \\
				% 		DexNet       & 0.977 & 0.899 & +0.078  & 2.76  & \textbf{0.006} \\
				% 		ISIC         & 0.923 & 0.823 & +0.100  & 3.87  & \textbf{0.0012} \\
				% 		PlantVillage & 0.914 & 0.860 & +0.054  & 3.05  & \textbf{0.0030} \\
				% 		\bottomrule
				% 	\end{tabular}
			% \end{table}

		\begin{table}[t]
			\centering
			\caption{\textbf{Caption entropy analysis (H condition).}
				Entropy compared between correct and wrong predictions. Values show mean entropy for each group, their difference, and two-sided $t$-tests.
				Significant results ($p<0.05$) are bolded.}
			\label{tab:entropy_analysis}
			\resizebox{0.7\textwidth}{!}{%
				{\renewcommand{\arraystretch}{1.3}
					\begin{tabular}{lccccc}
						\toprule
						\rowcolor{headergray}
						\textbf{Dataset} & \textbf{Correct} & \textbf{Wrong} & \boldmath$\Delta$\unboldmath (C{-}W) & \textbf{$t$-stat} & \textbf{$p$-value} \\
						\midrule
						\rowcolor{gray!10}
						AntiCP2      & $0.861$ & $0.856$ & $+0.005$  & $0.32$  & $0.749$ \\
						BBBP         & $0.886$ & $0.960$ & $-0.074$  & $-2.26$ & \textbf{$0.032$} \\
						\rowcolor{gray!10}
						CodeXGLUE    & $0.901$ & $0.897$ & $+0.004$  & $0.30$  & $0.768$ \\
						SARA\_V3     & $1.030$ & $1.004$ & $+0.025$  & $1.91$  & $0.060$ \\
						\rowcolor{gray!10}
						ProofWriter  & $0.975$ & $0.998$ & $-0.023$  & $-1.14$ & $0.262$ \\
						GQA          & $1.150$ & $1.048$ & $+0.102$  & $4.61$  & \textbf{$1.4{\times}10^{-5}$} \\
						\rowcolor{gray!10}
						DexNet       & $0.977$ & $0.899$ & $+0.078$  & $2.76$  & \bm{$0.006$} \\
						ISIC         & $0.923$ & $0.823$ & $+0.100$  & $3.87$  & \bm{$0.0012$} \\
						\rowcolor{gray!10}
						PlantVillage & $0.914$ & $0.860$ & $+0.054$  & $3.05$  & \bm{$0.0030$} \\
						\bottomrule
				\end{tabular}}
			}
		\end{table}

		\section{Implementation Details of Convergence and Similarity Analysis}
		\label{app:analysis_convergence}
			\hspace*{2em}
		\textbf{Intra-chain convergence.} To further understand the internal dynamics of hallucinated reasoning, 
		we analyze whether intermediate steps in a reasoning chain progressively converge
		toward the final conclusion. Specifically, we extract step-wise reasoning traces from 
		hallucinated captions and compute semantic embeddings using OpenCLIP (ViT-L/14, OpenAI weights). 
		Each intermediate step is compared to the final step via cosine similarity, yielding a 
		\emph{step-to-final similarity curve} averaged across reasoning chains. Similarity consistently increases 
		as the chain progresses, while variance bands narrow, indicating that hallucinated reasoning 
		exhibits stable intra-chain convergence. This suggests that intermediate steps are not drifting 
		away but instead steadily aligning with the final conclusion.
		
		\textbf{Inter-chain convergence.} To further evaluate the stability of reasoning trajectories, 
		we computed the \emph{average path similarity} across multiple sampled chains. 
		For each dataset, we first extracted hallucinated (H) and non-hallucinated (NH) reasoning paths, 
		then embedded all intermediate steps using OpenCLIP (ViT-L/14, OpenAI weights). 
		The cosine similarity between different runs was averaged to yield an overall 
		\emph{path-level similarity score}. We then compared the distribution of average similarities between H and NH conditions. 
		Kernel density estimation (KDE) was applied to visualize the distributions. Results indicate that both H and NH paths 
		consistently achieve very high similarity (means $\approx 0.97$), with nearly overlapping distributions. 
		This confirms that hallucinations do not compromise inter-chain stability, and that 
		multiple reasoning paths remain semantically aligned across runs. 
		
%		\section{HIVE performance experiment}
%		\label{Method_performance}
%		To assess the reliability of HIVE’s hallucination discriminator, we evaluate it in two settings. First, on the TruthfulQA benchmark, which is commonly used to probe hallucination, the discriminator achieves 81.76\% accuracy. Second, to approximate real-world, cross-domain use, we curate a 180-sample dataset by sampling 20 captions from each of nine tasks, manually annotate them as either hallucination or faithful, and back-test the discriminator; it attains 83.72\% accuracy. These results demonstrate that the module generalizes beyond a single benchmark and effectively separates hallucinated from faithful captions, providing a stable foundation for subsequent comparisons.

\section{Additional Robustness Experiments}
\label{sec:additional_robustness}
\hspace*{2em}
To test whether the observed PHR effect is limited to our original domain-specific datasets, we further evaluate GPT-4o on two general multimodal benchmarks, MMStar and MMBench. We focus on perception and reasoning subsets that align with the scope of our study. As shown in Table~\ref{tab:general_benchmark_gains}, hallucinated captions yield directionally consistent gains across all evaluated subsets.

\begin{table}[t]
	\centering
	\scriptsize
	\renewcommand{\arraystretch}{1.06}
	\setlength{\tabcolsep}{5pt}
	\caption{\textbf{General benchmark gains.}
		GPT-4o results on MMStar and MMBench. Values report the accuracy gain of hallucinated captions over faithful captions.}
	\label{tab:general_benchmark_gains}
	\begin{tabular}{lcc}
		\toprule
		\rowcolor{headergray}
		\textbf{Subset} & \textbf{MMStar} & \textbf{MMBench} \\
		\midrule
		\rowcolor{rowgray}
		Coarse Perc.       & +3.13 & +5.77 \\
		Fine-grained Perc. & +4.17 & +0.38 \\
		\rowcolor{rowgray}
		Instance Reason.   & +6.25 & +2.14 \\
		Logical Reason.    & +4.29 & +4.44 \\
		\rowcolor{rowgray}
		\textbf{Avg.}      & \textbf{+4.46} & \textbf{+3.18} \\
		\bottomrule
	\end{tabular}
\end{table}

\section{License}
\label{Appendix:License}
	\hspace*{2em}
\textbf{Datasets.} All datasets used in this study are publicly available benchmarks.
Their license terms are as follows: AntiCP2 is released under GPL-3.0; BBBP under the MIT License; CodeXGLUE under the Computational Use of Data Agreement (C-UDA); SARA\_V3 under CC BY 4.0; ProofWriter under CC BY 4.0; GQA annotations under CC BY 4.0; Dex-Net code under BSD-3-Clause while its HDF5 databases are restricted to research-only (non-commercial) use; ISIC under CC BY-NC (non-commercial); and PlantVillage under CC0.
We emphasize that our use of these datasets is strictly for academic research purposes.

\textbf{Models.} All models used in this study are publicly available APIs or checkpoints released by their respective providers.
Specifically, Qwen2.5-VL-3B and Qwen2.5-VL-72B are released under the Qwen Research License, while Qwen2.5-VL-7B and Qwen2.5-VL-32B adopt the Apache 2.0 License.
For commercial API models, including GPT-4o, GPT-3.5-turbo (OpenAI), Claude 3 Sonnet (Anthropic), Gemini 2.0 Flash (Google DeepMind), O3 (OpenAI), DeepSeek-V3 and DeepSeek-R1 (DeepSeek), Mistral Large (Mistral), and Qwen-VL (Alibaba), usage is governed by their providers’ service terms and API agreements.
We emphasize that our use of these models is strictly for academic research purposes in accordance with their public availability and license terms.

\section{AI Disclosure}
\label{appendix:disclosure}
\hspace*{2em}
We acknowledge the use of GPT-5 for grammar checking only. The model was employed to correct grammatical errors while ensuring the original meaning and intent of the text remained unchanged.

\end{document}